\documentclass[letterpaper]{article} 
\usepackage[preprint]{aaai2027}  
\usepackage[hyphens]{url}  
\usepackage{graphicx} 
\usepackage{natbib}  
\usepackage{caption} 
\usepackage{amsmath}
\usepackage{amssymb}
\usepackage{xspace}
\newcommand{\method}{\textsc{AnchorFold}\xspace}
\usepackage{booktabs}
\usepackage{multirow}

\title{\method: A Focus-Then-Fold Framework via Recursive Attention Propagation for Efficient Multi-Vector Visual Document Retrieval}

\author{
    Haoyu Zuo\textsuperscript{\rm 1},
    Yibo Yan\textsuperscript{\rm 1,2,3},
    Xin Zou\textsuperscript{\rm 1,3},
    Shuliang Liu\textsuperscript{\rm 1,3},
    Yi Cao\textsuperscript{\rm 2},
    Mingdong Ou\textsuperscript{\rm 2,}\thanks{Project lead.},
    Xuming Hu\textsuperscript{\rm 1,3,}\thanks{Corresponding author.}
}
\affiliations{
    \textsuperscript{\rm 1}Hong Kong University of Science and Technology (Guangzhou)\\
    \textsuperscript{\rm 2}Alibaba Cloud Computing,
    \textsuperscript{\rm 3}Hong Kong University of Science and Technology\\
    \texttt{hzuo258@connect.hkust-gz.edu.cn, xuminghu@hkust-gz.edu.cn}
}

\begin{document}

\maketitle

\begin{abstract}
Multi-vector vision-language retrievers enable fine-grained Visual Document Retrieval (VDR) through late interaction, but storing and scoring hundreds of visual patch embeddings per page incurs substantial overhead. Existing training-free methods rely on pruning or merging: pruning degrades sharply under aggressive compression, whereas merging does not explicitly prioritize important regions when forming representatives. We introduce \textbf{\method, a training-free focus-then-fold framework for document-side index compression}. \method applies Recursive Attention Propagation over visual self-attention graphs, performing multi-step propagation within each attention head and integrating scores across heads and layers. The focus stage selects the highest-centrality tokens as anchors. The fold stage assigns remaining tokens to their most similar anchors in the normalized retrieval space and summarizes each anchor-centered group through centrality-weighted aggregation. This preserves non-anchor contributions while concentrating capacity on structurally important tokens. Across ViDoRe v1/v2 and REAL-MM-RAG with three diverse retrieval backbones, \method consistently outperforms all evaluated training-free baselines at $\gamma\leq0.20$. On ViDoRe v1/v2, it retains 98.3\% of full-index NDCG@5 on average at $5\times$ compression, achieving near-lossless compression, and 92.4\% at $20\times$ compression.
\end{abstract}

\section{Introduction}

\begin{figure}[t]
\centering
\includegraphics[width=\columnwidth]
{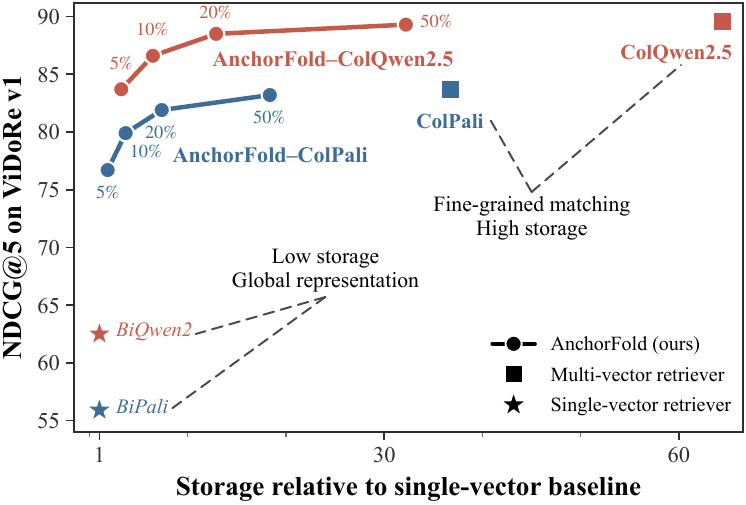}
\caption{
Retrieval effectiveness versus relative document-embedding storage on ViDoRe v1. Relative storage is normalized to the corresponding single-vector baseline; percentages denote vector retention ratios.
}
\label{fig:memory_tradeoff}
\end{figure}

Visual Document Retrieval (VDR) aims to retrieve pages relevant to textual queries from visually rich document corpora \citep{yan2026unlocking,tanaka2025vdocrag,wang2025vidorag}. Unlike text-only retrieval, relevance can depend jointly on textual content, tables, figures, and their spatial layout. Single-vector retrievers represent each page with one global embedding, whereas multi-vector vision-language retrievers such as ColPali encode each page into contextualized visual patch embeddings and use late interaction: MaxSim matches each query-token embedding to its most similar document vector \citep{ma2024dse,faysse2024colpali,khattab2020colbert}. By deferring aggregation until query-time scoring, late interaction preserves localized matching signals that a global representation may obscure. This fine-grained representation, however, incurs substantial storage and scoring costs: each page contributes hundreds of index vectors, and both index storage and exact per-document MaxSim cost scale linearly with the number of stored vectors \citep{santhanam2022colbertv2,nardini2024emvb}. Figure~\ref{fig:memory_tradeoff} illustrates this trade-off: \method retains most of the full-index retrieval effectiveness while requiring substantially less document-embedding storage. Reducing the document-side vector count is therefore critical to scaling multi-vector VDR.

\begin{figure*}[t]
\centering
\includegraphics[width=0.75\textwidth]{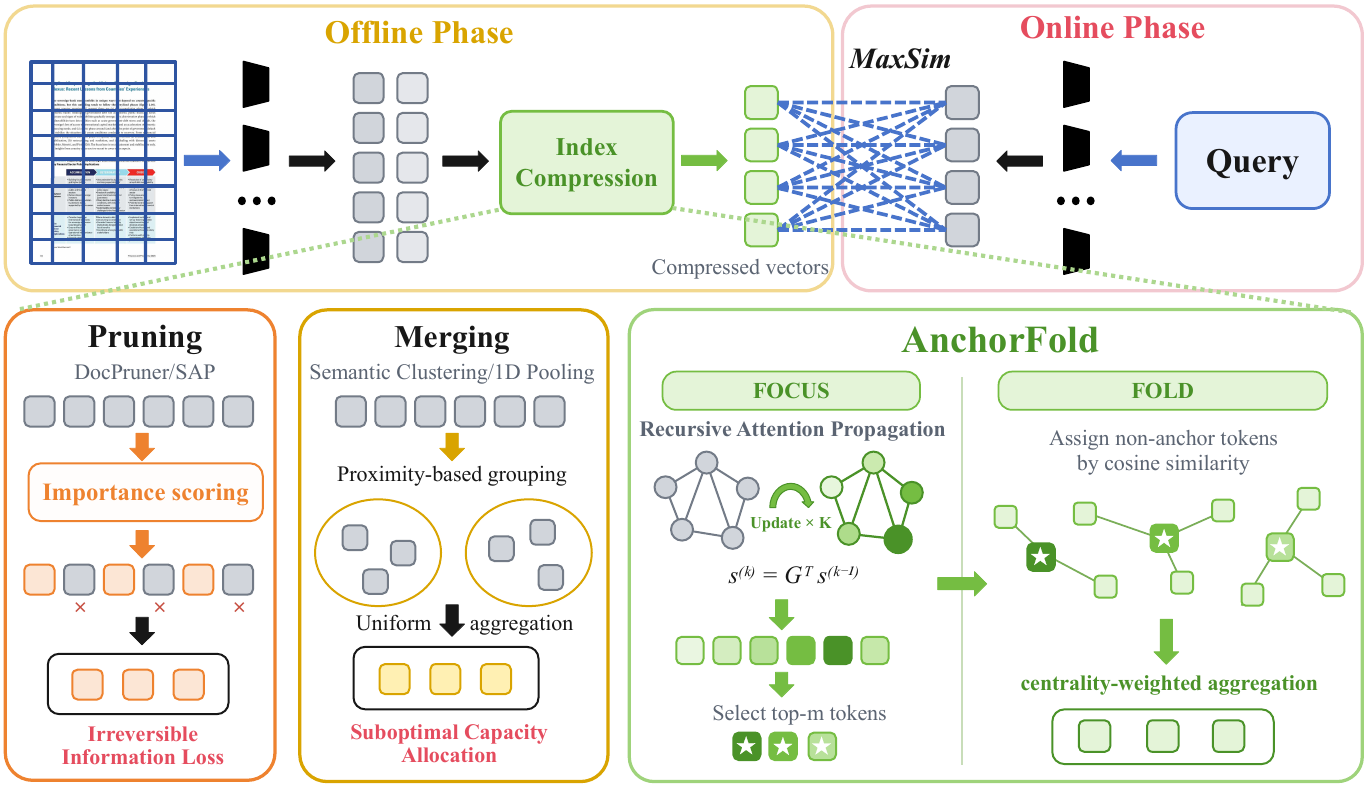}
\caption{
Overview of \method for training-free multi-vector VDR compression. Top: document representations are compressed during offline indexing and reused for online MaxSim retrieval. Bottom: pruning discards unselected vectors, while merging forms representatives through proximity-based grouping and uniform aggregation. \method instead follows a focus-then-fold process, selecting anchors via Recursive Attention Propagation and folding non-anchor tokens into anchor-centered representations through similarity-based assignment and centrality-weighted aggregation.
}
\label{fig:method_overview}
\end{figure*}

Existing training-free methods for document-side index compression, illustrated in the lower panel of Figure~\ref{fig:method_overview}, fall into two categories: pruning and merging. Pruning retains a subset of patch embeddings using attention-derived or structure-aware selection scores \citep{yan2025docpruner,liu2026sap}. It leaves selected vectors unchanged with relatively low compression-time overhead. However, unselected vectors are irreversibly removed. Because queries are unavailable during indexing, a discarded vector may still provide a strong MaxSim match for a future query token \citep{zong2025losslesspruning}. Under aggressive pruning, this loss of candidate matches can degrade retrieval effectiveness \citep{yan2026prunethenmerge}. Merging instead replaces groups of patch embeddings with fewer representative vectors through spatial pooling or representation-space clustering \citep{ma2025lightcolpali}. By allowing a broader set of tokens to influence the compressed representation, merging can degrade more gracefully as the document-vector budget decreases. However, purely spatial or similarity-based grouping does not explicitly allocate limited representation capacity according to the importance of page regions, while uniform aggregation can attenuate localized discriminative features \citep{veneroso2025crisp}. Learned compressors can optimize representative selection and aggregation, but require compression-specific training or learnable modules \citep{xiao2025metaembed,huo2026causalembed}.

These limitations expose a central trade-off under a constrained document-vector budget: a compressor must construct a small set of representative vectors while preserving evidence distributed across the page. Effective training-free compression must therefore address two coupled questions: \textit{where should the limited representation capacity be allocated}, and \textit{how can tokens not stored individually still influence the compressed representation?}

We introduce \textbf{\method, a training-free framework for document-side index compression that organizes compression into a focus-then-fold process}. \method operates during offline index construction, introduces no trainable parameters, and leaves both query encoding and late-interaction scoring unchanged. In the focus stage, \method constructs visual-to-visual self-attention graphs from a fixed window of intermediate backbone layers. It then performs finite-step Recursive Attention Propagation independently within each attention head, allowing support to be redistributed along multi-hop attention paths. Integrating the resulting signals across heads and layers yields query-agnostic propagated attention centrality scores, which are used to select the highest-scoring tokens as anchors. In the fold stage, each non-anchor token is assigned to its most similar anchor in the normalized retrieval space, and each anchor-centered group is summarized via centrality-weighted aggregation. Propagated attention centrality allocates representation capacity, while retrieval-space similarity guides token assignment. This design retains contributions from non-anchor tokens while concentrating the available capacity on tokens receiving stronger multi-hop attention support.

We evaluate \method on ViDoRe v1 and v2 \citep{faysse2024colpali,mace2025vidorev2} using three heterogeneous multi-vector retrieval backbones, and further assess its generalization to REAL-MM-RAG \citep{wasserman2025realmmrag}. Across these benchmarks and all three backbones, \method consistently outperforms all evaluated training-free compression baselines at retention ratios of $20\%$, $10\%$, and $5\%$. On ViDoRe v1/v2, it retains 98.3\% of full-index NDCG@5 on average at $5\times$ compression, achieving near-lossless compression; even at $20\times$ compression, it preserves 92.4\%. The widening performance margin under stronger compression, also observed on REAL-MM-RAG, suggests that concentrating the vector budget on high-centrality anchors while consolidating evidence from non-anchor tokens becomes increasingly beneficial as the document-vector budget decreases.

\section{Related Work}
\label{sec:related_work}

\subsection{Multi-Vector Retrieval}

ColBERT introduced late interaction in text retrieval, representing documents as sets of contextualized token embeddings for fine-grained query--document matching \citep{khattab2020colbert}. ColPali extended this paradigm to VDR by encoding document images into multi-vector representations \citep{faysse2024colpali}. Subsequent work advances VDR through vision-language backbones \citep{bai2025qwen25vl} and retrieval-oriented architectures and training strategies \citep{gunther2025jinav4,moreira2026nemotron,chaffin2026colbert}. Although these representations preserve localized matching signals, they associate each page with many vectors, making index efficiency important at scale.

\subsection{Efficient Multi-Vector VDR}

\paragraph{Pruning-based compression.}
Pruning retains a subset of visual embeddings \citep{acquavia2023staticpruning,lassance2022learnedtokenpruning,zong2025losslesspruning,jha2026brief,archish2026incorporating}. Light-ColPali studies random and attention-based token selection \citep{ma2025lightcolpali}, while DocPruner uses intra-document attention statistics and SAP uses intermediate-layer visual in-degree centrality \citep{yan2025docpruner,liu2026sap}. These methods leave retained embeddings unchanged, but removed tokens no longer contribute to late-interaction matching.

\paragraph{Merging-based compression.}
Merging consolidates embeddings into fewer representatives \citep{clavie2024tokenpooling}. Light-ColPali studies spatial pooling and representation-space clustering \citep{ma2025lightcolpali}. Prune-then-Merge combines adaptive pruning with hierarchical clustering \citep{yan2026prunethenmerge}, while ColChunk augments hierarchical clustering with two-dimensional positional information \citep{yan2026colchunk}. However, grouping based primarily on spatial or representation-space proximity may not allocate distinct representation capacity to important page regions.

\paragraph{Learned and structure-assisted compact representations.}
Other work constructs compact representations through learned mechanisms or auxiliary structure \citep{macavaney2025constantspace,xiang2026mm,cha2026reinpool,jaasaari2026lemur}. MetaEmbed learns meta tokens with a Matryoshka multi-vector objective \citep{xiao2025metaembed}, while CausalEmbed autoregressively generates latent document representations \citep{huo2026causalembed}. AGC learns universal query tokens for saliency-guided centroid selection and weighted aggregation \citep{qin2026agc}, whereas ColParse uses parsed structure to build layout-aware representations \citep{yan2026colparse}. These approaches require training, learnable components, or preprocessing.

Training-free methods typically use importance signals for token retention or spatial and representation-space proximity for grouping. \method instead derives self-attention-based, query-agnostic importance for anchor selection and weighted aggregation. Similarity-based assignment allows non-anchor tokens to contribute to the compressed representation without modifying the query encoder or the original late-interaction scoring function.

\section{Methodology}

\subsection{Overview}

Figure~\ref{fig:method_overview} illustrates \method, a training-free framework that compresses document representations during offline index construction. It introduces no trainable parameters and leaves query encoding and MaxSim scoring unchanged.

Let $\mathbf{Z}=[\mathbf{z}_1,\ldots,\mathbf{z}_N]$ denote the final-layer hidden states of the $N$ valid visual tokens immediately before the retrieval projection, where $\mathbf{z}_i\in\mathbb{R}^{d}$. Let $P:\mathbb{R}^{d}\rightarrow\mathbb{R}^{d_r}$ denote the retrieval projection of the underlying backbone, with $P$ taken as the identity when the backbone already outputs retrieval-space representations. The uncompressed document index is $\mathbf{E}=[\mathbf{e}_1,\ldots,\mathbf{e}_N]$, where
\begin{equation}
\mathbf{e}_i
=
\operatorname{norm}(P(\mathbf{z}_i)),
\qquad
\operatorname{norm}(\mathbf{u})
=
\frac{\mathbf{u}}{\lVert \mathbf{u}\rVert_2}.
\end{equation}

Given a target retention ratio $\gamma\in(0,1]$, we set the compressed document-vector budget to
\begin{equation}
m
=
\left\lceil \gamma N \right\rceil.
\end{equation}

\method follows a focus-then-fold process. The focus stage estimates propagated attention centrality and selects the $m$ highest-scoring tokens as anchors, thereby allocating the available representation budget. The fold stage assigns each non-anchor token to its most similar anchor in the normalized retrieval space and summarizes each anchor-centered group through centrality-weighted aggregation.

\subsection{Recursive Attention Propagation}

We derive an attention-based centrality signal from the visual self-attention patterns of the retrieval backbone \citep{wang2023zerotprune}. For a Transformer layer $l$ and attention head $h$, let
$\mathbf{A}^{l,h}\in[0,1]^{T\times T}$
denote the post-softmax self-attention matrix, where $\mathbf{A}^{l,h}_{ab}$ is the attention weight from query position $a$ to key position $b$, and let
$V=\{v_1,\ldots,v_N\}$
denote the positions of the valid visual tokens in the full sequence. We form a visual attention graph by restricting $\mathbf{A}^{l,h}$ to these tokens and row-normalizing the resulting submatrix:
\begin{equation}
\mathbf{G}^{l,h}_{ij}
=
\frac{
\mathbf{A}^{l,h}_{v_i v_j}
}{
\sum_{r=1}^{N}
\mathbf{A}^{l,h}_{v_i v_r}
}.
\end{equation}
$\mathbf{G}^{l,h}\in\mathbb{R}^{N\times N}$ is row-stochastic and defines a directed visual attention transition matrix for head $h$ at layer $l$.

One-step aggregation captures only first-order support because every token initially contributes equal importance mass. To incorporate multi-hop attention structure, we initialize a uniform distribution over the valid visual tokens and propagate it for $K$ steps:
\begin{equation}
\mathbf{s}^{l,h}_{0}
=
\frac{1}{N}\mathbf{1},
\qquad
\mathbf{s}^{l,h}_{t}
=
\left(\mathbf{G}^{l,h}\right)^{\top}
\mathbf{s}^{l,h}_{t-1},
\quad
t=1,\ldots,K.
\end{equation}
We term this procedure Recursive Attention Propagation. After $K$ steps, a token receives a high score when it is supported by tokens that have themselves accumulated importance, thereby incorporating multi-hop attention structure.

Propagation is performed independently within each attention head. We then integrate the resulting head-specific scores using a root-mean-square operator, which emphasizes stronger head-specific responses while integrating evidence across all heads:
\begin{equation}
r^{l}(j)
=
\sqrt{
\frac{1}{H}
\sum_{h=1}^{H}
\left(
\mathbf{s}^{l,h}_{K}(j)
\right)^2
}.
\end{equation}
To remove layer-wise scale differences before averaging, we normalize each layer score by its mean over the valid visual tokens:
\begin{equation}
\tilde{r}^{l}(j)
=
\frac{
r^{l}(j)
}{
\frac{1}{N}
\sum_{q=1}^{N}
r^{l}(q)
}.
\end{equation}

Because attention organization varies across backbone depth, centrality derived from a single layer can be sensitive to layer-specific attention patterns \citep{liu2026sap}. We therefore integrate the normalized scores over a contiguous layer set $\mathcal{L}$. The selection of $\mathcal{L}$ is described in Section~\ref{sec:hyperparameter_analysis}. The final propagated attention centrality score of token $j$ is defined as
\begin{equation}
\alpha_j
=
\frac{1}{|\mathcal{L}|}
\sum_{l\in\mathcal{L}}
\tilde{r}^{l}(j).
\end{equation}
The resulting centrality scores
$\boldsymbol{\alpha}=[\alpha_1,\ldots,\alpha_N]$
are used to allocate the anchor budget and to modulate the contribution of tokens during subsequent aggregation.

\subsection{Anchor Selection}

Given the target vector budget $m$, the focus stage selects the indices of the $m$ tokens with the largest propagated attention centrality scores:
\begin{equation}
C
=
\{c_1,\ldots,c_m\}
=
\operatorname{Top}_{m}
\left(
\boldsymbol{\alpha}
\right),
\end{equation}
where
$C\subseteq\{1,\ldots,N\}$.
The selected token indices define the anchors used in the subsequent fold stage.

\subsection{Token Assignment and Weighted Aggregation}

The fold stage begins by assigning each non-anchor token to the most similar anchor using the L2-normalized retrieval vectors $\mathbf{e}_i$:
\begin{equation} a(i) = \arg\max_{1\leq q\leq m} \mathbf{e}_i^{\top}\mathbf{e}_{c_q}, \qquad i\notin C. \end{equation}
This cosine-based assignment is performed in the normalized retrieval space used by the downstream MaxSim scoring function.

For each anchor $c_q$, we define the corresponding anchor-centered group as
\begin{equation}
\mathcal{C}_q
=
\{c_q\}
\cup
\left\{
i\notin C
\mid
a(i)=q
\right\}.
\end{equation}
This construction guarantees that every anchor remains in its own group and produces exactly $m$ non-empty groups. When $m=N$, \method reduces to the original uncompressed document representation.

The fold stage is completed by consolidating each anchor-centered group into a single document vector. For each group $\mathcal{C}_q$, we aggregate the pre-projection hidden states $\mathbf{z}_i$ using the propagated attention centrality scores:
\begin{equation}
\mathbf{y}_q
=
\frac{
\sum_{i\in\mathcal{C}_q}
\alpha_i \mathbf{z}_i
}{
\sum_{i\in\mathcal{C}_q}
\alpha_i
}.
\end{equation}
Compared with uniform averaging, centrality-weighted aggregation reduces the relative contribution of tokens with lower propagated attention centrality while still allowing every token assigned to the group to contribute to the resulting representation. The compressed document index is
\begin{equation}
\widehat{\mathbf{E}}
=
[\hat{\mathbf{e}}_1,\ldots,\hat{\mathbf{e}}_m],
\qquad
\hat{\mathbf{e}}_q
=
\operatorname{norm}\left(P(\mathbf{y}_q)\right).
\end{equation}
Retrieval is then performed using the original MaxSim scoring function over the compressed document index $\widehat{\mathbf{E}}$.

\begin{figure*}[t]
    \centering
    \includegraphics[width=0.80\textwidth]
    {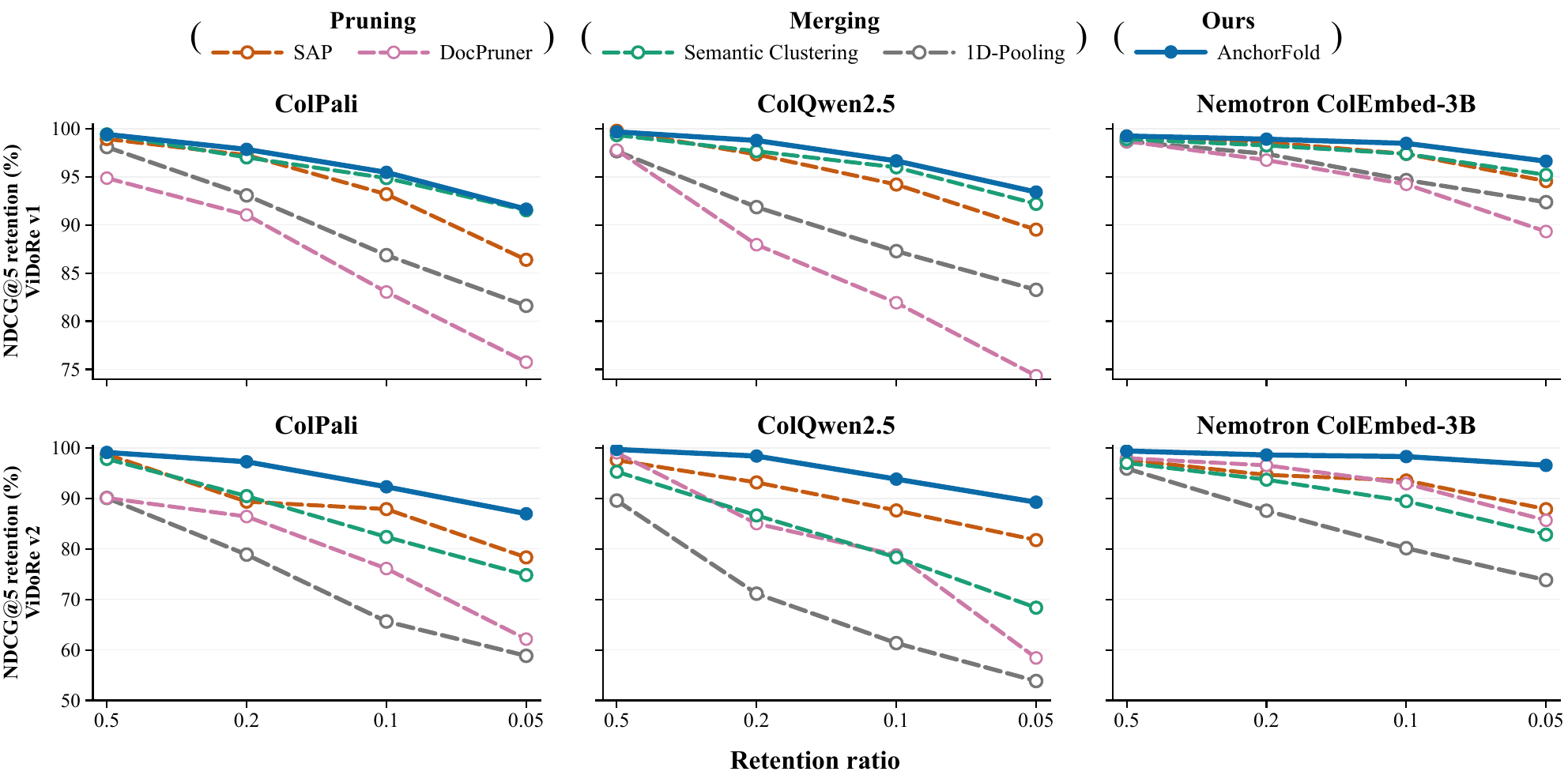}
    \caption{
    NDCG@5 retention on ViDoRe v1 (top) and v2 (bottom) across three retrieval backbones. \method achieves the highest retention among the evaluated methods in all six settings at $\gamma\leq0.20$, with larger margins under stronger compression.
    }
    \label{fig:main_results}
\end{figure*}

\section{Experiments}

\subsection{Experimental Setup}

\paragraph{Benchmarks and evaluation.}
We conduct experiments on the full ViDoRe v1 benchmark and the four datasets in the official ViDoRe v2 collection, following their standard evaluation protocols \citep{faysse2024colpali,mace2025vidorev2}. ViDoRe v1 comprises ten datasets spanning diverse document domains and formats, while ViDoRe v2 comprises four more challenging datasets covering ESG reports, biomedical lectures, and economics reports. We further evaluate generalization on the BEIR-compatible versions of the four REAL-MM-RAG datasets: FinReport, FinSlides, TechReport, and TechSlides \citep{wasserman2025realmmrag}. Following standard VDR practice, we adopt NDCG@5 as the primary evaluation metric \citep{wang2013ndcg} and report the unweighted macro-average across the constituent datasets of each benchmark. We additionally report NDCG@5 retention, defined as the benchmark-level NDCG@5 of a compressed index divided by that of its corresponding full index.

\paragraph{Retrieval backbones.}
We evaluate \method with three representative multi-vector VDR backbones: ColPali v1.3 \citep{faysse2024colpali}, ColQwen2.5 with bidirectional attention \citep{bai2025qwen25vl}, and Nemotron ColEmbed-3B v2 \citep{moreira2026nemotron}. The three models contain 18, 36, and 28 backbone layers, respectively, and differ substantially in backbone architecture, visual tokenization, and model depth. All parameters of the underlying retrievers remain frozen, and \method is applied exclusively to document representations during offline index construction.

\paragraph{Baselines.}
We include the uncompressed Full Index as the reference and compare \method with four representative training-free compression baselines: SAP, which performs structure-aware pruning using visual in-degree centrality \citep{liu2026sap}; Semantic Clustering, which groups final document embeddings based on representation-space proximity and stores the resulting centroids \citep{ma2025lightcolpali}; 1D-Pooling, which averages consecutive visual embeddings \citep{ma2025lightcolpali}; and DocPruner, an adaptive attention-based pruning method \citep{yan2025docpruner}. We evaluate retention ratios $\gamma\in\{0.50,0.20,0.10,0.05\}$; for fixed-ratio methods, a page with $N$ valid visual tokens is compressed to $\lceil\gamma N\rceil$ vectors. To fairly compare with the fixed-ratio methods, we apply quantile-based calibration to DocPruner on a held-out calibration set, aligning its global retention rate with each target retention ratio \citep{liu2026sap}.

\paragraph{Implementation details.}
All compression methods are implemented within a unified PyTorch evaluation pipeline. Within each backbone, all methods are evaluated on the same document inputs at matched target retention ratios. \method uses $K=6$ with layer windows 11--14, 25--32, and 14--19 for ColPali, ColQwen2.5, and Nemotron ColEmbed-3B, respectively, as selected by the SR-based calibration in Section~\ref{sec:hyperparameter_analysis}. All experiments are conducted on a cluster equipped with NVIDIA A800 GPUs. Additional dataset statistics, model specifications, baseline settings, and implementation details are provided in Appendix A.

\subsection{Main Results}

Figure~\ref{fig:main_results} compares NDCG@5 retention across three retrieval backbones on ViDoRe v1 and v2, revealing three consistent patterns across different backbone--benchmark pairs.

First, under moderate compression, \method preserves nearly all of the full-index retrieval effectiveness. At $\gamma=0.20$, corresponding to a roughly $5\times$ reduction in the number of stored document vectors, \method achieves the highest retention on all six backbone--benchmark pairs, with a mean NDCG@5 retention of 98.3\% across the six pairs. At $\gamma=0.50$, the performance gaps among the compression strategies are comparatively small.

Second, the advantage of \method becomes increasingly pronounced as the representation budget decreases. At $\gamma=0.10$ and $\gamma=0.05$, \method retains 95.8\% and 92.4\% of full-index NDCG@5 on average, respectively, while achieving the best performance on all six backbone--benchmark pairs. Relative to the strongest evaluated baseline for each pair, the mean margin in NDCG@5 retention increases from 2.7 percentage points at $\gamma=0.20$ to 3.0 points at $\gamma=0.10$ and 4.6 points at $\gamma=0.05$. This trend indicates that allocating the limited vector budget to high-centrality anchors while folding non-anchor evidence into the resulting compressed representations becomes particularly beneficial under aggressive compression.

Third, the gains are more pronounced on the more challenging ViDoRe v2 benchmark and remain consistent across heterogeneous retrieval backbones. At $\gamma=0.05$, \method exceeds the strongest evaluated baseline by 4.7, 4.6, and 5.5 NDCG@5 points on ViDoRe v2 for ColPali, ColQwen2.5, and Nemotron ColEmbed-3B, respectively, compared with gains of 0.1, 1.1, and 1.3 points on ViDoRe v1. Notably, \method retains 96.5\% of full-index NDCG@5 with Nemotron ColEmbed-3B on ViDoRe v2 at $\gamma=0.05$. Taken together, the consistent improvements across three substantially different retrieval backbones indicate that the effectiveness of \method is not tied to a particular architecture. Its larger margins on ViDoRe v2 suggest a more favorable degradation profile when the document-vector budget is severely constrained. Complete NDCG@5 and retention results are provided in Appendix B.

\begin{figure*}[t]
    \centering
    \includegraphics[width=0.85\textwidth]
    {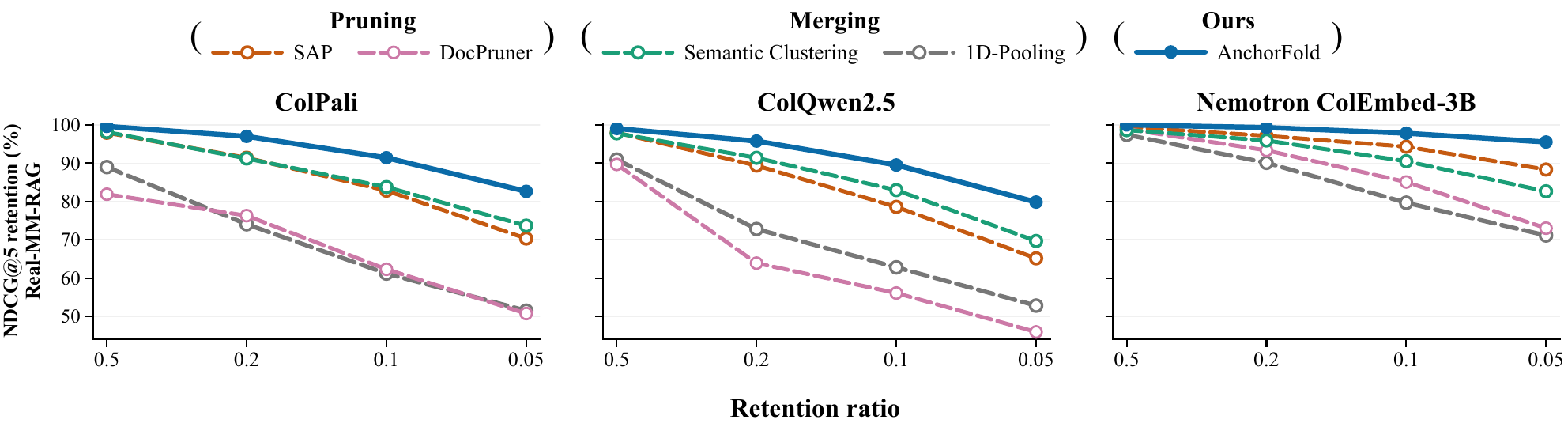}
    \caption{
    NDCG@5 retention on REAL-MM-RAG across three retrieval backbones. \method achieves the highest retention among the evaluated methods, with larger margins under stronger compression.
    }
    \label{fig:real_mm_rag}
\end{figure*}

\subsection{Generalization to Complex Retrieval Settings}
\label{sec:real_mm_rag}

We further evaluate \method on REAL-MM-RAG, which uses rephrased queries to assess retrieval robustness under reduced lexical overlap between queries and relevant documents \citep{wasserman2025realmmrag}. As shown in Figure~\ref{fig:real_mm_rag}, \method consistently achieves the highest NDCG@5 retention among the evaluated methods across all three backbones, with its advantage becoming more pronounced as the retention ratio decreases. At $\gamma=0.20$, \method retains 97.4\% of full-index NDCG@5 on average and exceeds the strongest evaluated baseline by 4.0 percentage points.

The advantage further increases under aggressive compression. At $\gamma=0.05$, \method outperforms the strongest evaluated baseline by 4.8, 6.5, and 5.3 NDCG@5 points for ColPali, ColQwen2.5, and Nemotron ColEmbed-3B, respectively. These results show that the advantages observed on ViDoRe extend to the evaluated REAL-MM-RAG setting and remain substantial when the available document-vector budget is severely constrained. Complete per-dataset results are provided in Appendix B.

\subsection{Comparison with Trained Compression}

We compare \method with AGC, a trained compression method based on learned universal query tokens \citep{qin2026agc}. For a controlled comparison, we reproduce AGC within a unified codebase using the same ColQwen2.5 configuration and training recipe as the baseline retriever, while \method is applied to the baseline checkpoint without additional training. Both are evaluated on ViDoRe v1, ViDoRe v2, and REAL-MM-RAG with a fixed budget of 128 document vectors. \method achieves an NDCG@5 of 59.8 on ViDoRe v2, exceeding the 57.0 obtained by AGC under the same setting, with consistent improvements also observed on ViDoRe v1 and REAL-MM-RAG. These results suggest that \method remains competitive with trained compression without requiring compression-specific training. The controlled comparison protocol and complete results are provided in Appendix C.

\begin{table}[t]
    \centering
    {\small
    \begin{tabular*}{0.95\columnwidth}
        {@{\extracolsep{\fill}}lcc@{}}
        \toprule
        Variant
        & NDCG@5 $\uparrow$
        & $\Delta$ \\
        \midrule
        \textbf{\method}
            & \textbf{60.31}
            & -- \\
        w/o Multi-Step RAP
            & 58.53
            & $-1.78$ \\
        w/o Centrality Selection
            & 57.11
            & $-3.20$ \\
        w/o Folding
            & 57.79
            & $-2.52$ \\
        w/o Centrality Weighting
            & 59.48
            & $-0.83$ \\
        \bottomrule
    \end{tabular*}
    }
    \caption{
    Ablation results on ViDoRe v2 with ColQwen2.5 at $\gamma=0.20$. Each variant independently modifies one design choice of the full \method configuration.
    }
    \label{tab:ablation}
\end{table}

\subsection{Ablation Study}

We conduct an ablation study on ColQwen2.5 over ViDoRe v2 at $\gamma=0.20$
to analyze the contribution of each design choice. Specifically, w/o Multi-Step
RAP sets $K=1$; w/o Centrality Selection replaces centrality-based anchor
selection with uniform random selection; w/o Folding retains only the selected
anchor embeddings without assigning or aggregating non-anchor tokens; and w/o
Centrality Weighting replaces centrality-weighted aggregation with uniform
averaging within each anchor-centered group. As shown in
Table~\ref{tab:ablation}, removing centrality-based anchor selection causes the
largest degradation, demonstrating its importance for allocating the limited
representation budget. Removing folding and multi-step RAP also leads to clear
performance drops, validating the benefits of preserving non-anchor evidence
and capturing multi-hop attention structure. Centrality-weighted aggregation
provides additional but smaller improvements over uniform averaging.

\subsection{Hyperparameter Analysis}
\label{sec:hyperparameter_analysis}

\begin{figure}[t]
    \centering
    \includegraphics[width=1\columnwidth]
    {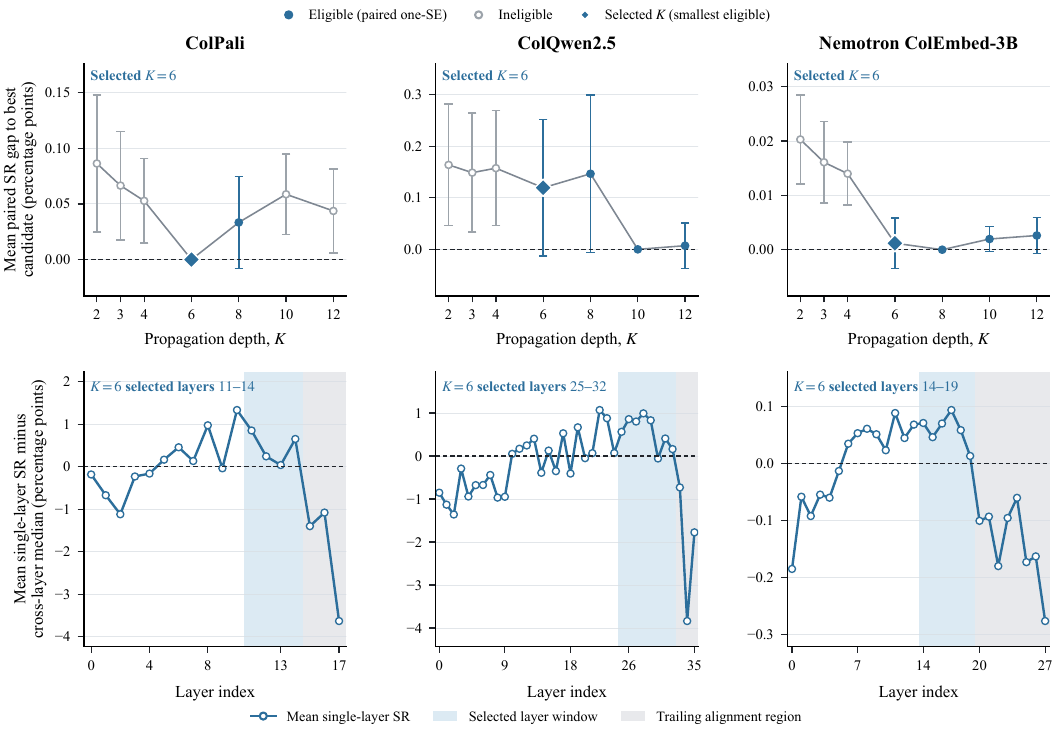}
    \caption{
    SR-based calibration of propagation depth and layer windows. Top: paired SR gaps across candidate depths. Bottom: median-centered layer-wise SR profiles at $K=6$, with selected windows and trailing low-SR regions shaded.
    }
    \label{fig:hyperparameter_analysis}
\end{figure}

\method uses a finite propagation depth $K$ together with a backbone-specific layer window. Following SAP, we jointly calibrate both choices once per backbone using SR, the compressed-to-full MaxSim score ratio, on 500 held-out query--page pairs sampled from the ColPali training corpus and disjoint from all evaluation splits, with a calibration retention ratio of $\gamma_{\mathrm{cal}}=0.20$ and a relative window width of $\rho=0.2$, corresponding to $20\%$ of the backbone depth \citep{liu2026sap}. For each candidate depth $K\in\{2,3,4,6,8,10,12\}$, we compute the mean single-layer SR profile, identify its longest trailing suffix below the across-layer median, place a candidate window immediately before this region, and evaluate the corresponding multi-layer \method operator. We select the smallest $K$ whose mean paired SR gap to the highest-mean joint configuration does not exceed one standard error of the per-pair gaps. Full calibration details and validation are provided in Appendix D.

As shown in Figure~\ref{fig:hyperparameter_analysis}, although the SR-maximizing joint configuration varies across backbones, the paired one-SE rule selects $K=6$ for all three models. All smaller candidates fall outside the eligible set, whereas larger depths yield only marginal or backbone-dependent improvements, making $K=6$ the most parsimonious configuration within the one-SE set. At $K=6$, the SR-guided procedure yields the zero-indexed layer windows 11--14, 25--32, and 14--19 for ColPali, ColQwen2.5, and Nemotron ColEmbed-3B, respectively. We therefore fix $K=6$ and these backbone-specific windows for all reported experiments.

\subsection{Efficiency Analysis}
\label{sec:efficiency}

We further evaluate the trade-off between retrieval fidelity and index-time efficiency on ViDoRe v2 using ColQwen2.5 at $\gamma=0.10$. Detailed efficiency evaluation settings and results are provided in Appendix E.

\begin{figure}[t]
    \centering
    \includegraphics[width=0.8\columnwidth]
    {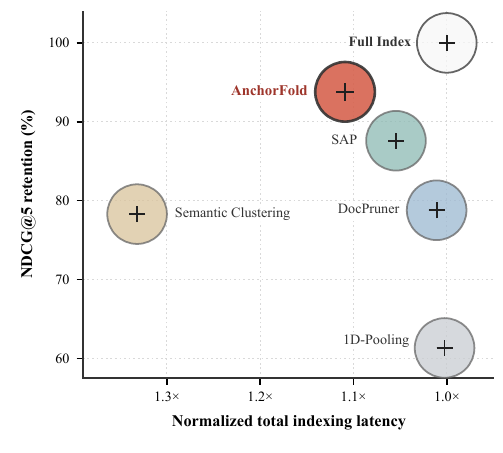}
    \caption{
    Retrieval fidelity--indexing latency trade-off on ViDoRe v2 with ColQwen2.5 at $\gamma=0.10$. Total indexing latency includes document encoding and compression and is normalized to the Full Index.
    }
    \label{fig:efficiency_tradeoff}
\end{figure}

As shown in Figure~\ref{fig:efficiency_tradeoff}, \method achieves the highest retrieval fidelity among the evaluated compression methods, retaining 93.80\% of full-index NDCG@5 at approximately $1.11\times$ the total indexing latency of the Full Index. Compared with SAP, \method provides substantially higher retrieval fidelity with a moderate increase in indexing cost, while outperforming Semantic Clustering in both retrieval fidelity and total indexing latency. By contrast, DocPruner and 1D-Pooling remain close to the Full Index in indexing latency but incur markedly larger losses in retrieval fidelity. Overall, \method offers a favorable trade-off between retrieval fidelity and index-time efficiency. Although it increases total indexing latency, the additional overhead is incurred exclusively during offline index construction and does not recur during subsequent queries.

\subsection{Qualitative Case Study}
\label{sec:case_study}

\begin{figure}[t]
    \centering
    \includegraphics[width=\columnwidth]
    {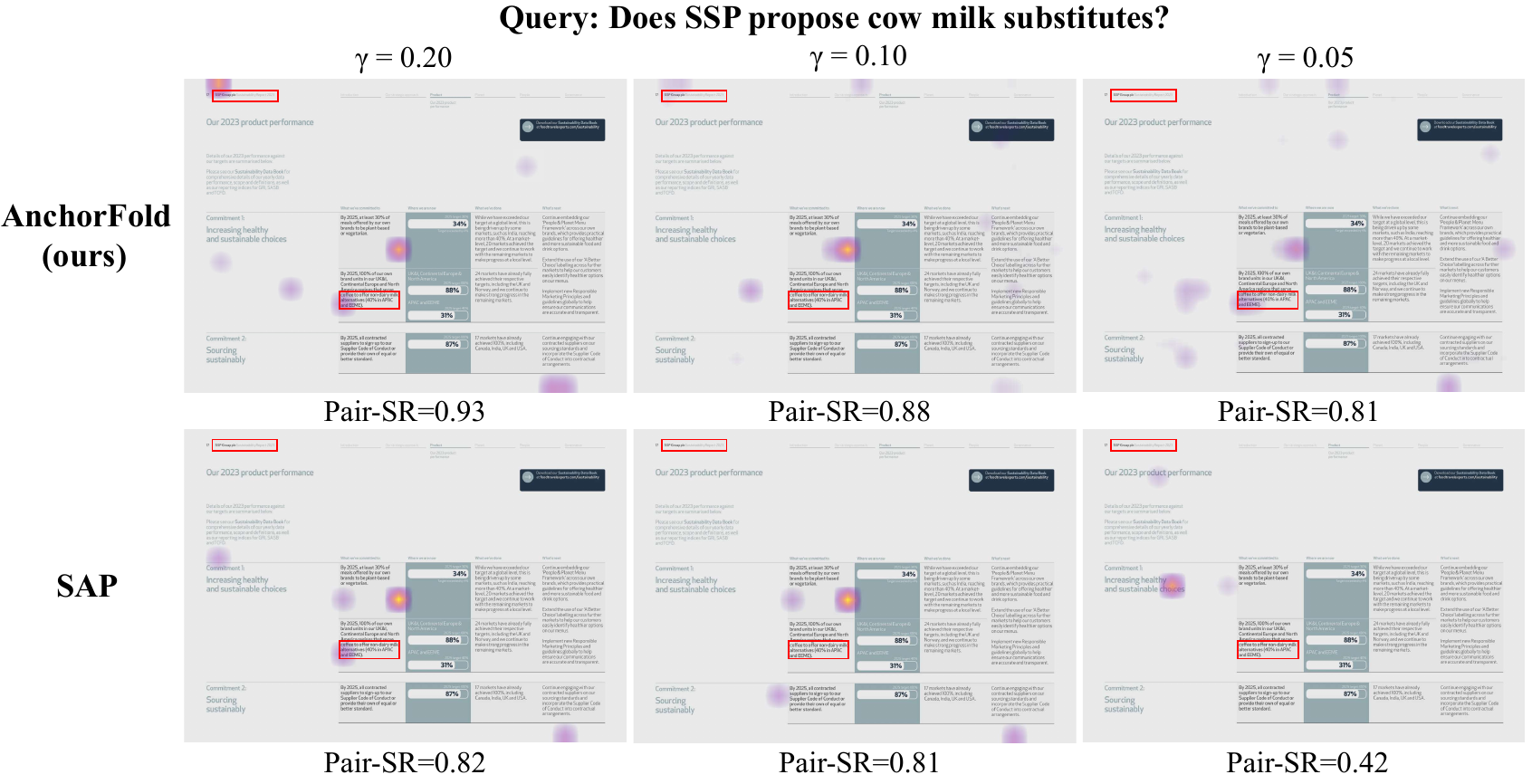}
    \caption{
    Qualitative comparison of \method and SAP on a ViDoRe v2 ESG query--page pair. Columns correspond to retention ratios $\gamma\in\{0.20,0.10,0.05\}$. The heatmaps show positive MaxSim contributions on a shared scale, and the red boxes delineate two query-relevant evidence regions. Pair-SR reports SR for the displayed query--page pair.
    }
    \label{fig:pair_case_study}
\end{figure}

Figure~\ref{fig:pair_case_study} compares how \method and SAP preserve
query-relevant matching signals as the document-vector budget decreases.
Across all three retention ratios, \method consistently yields a higher
Pair-SR than SAP, indicating more faithful preservation of the original
query--page matching score under the same vector budget. At
$\gamma=0.20$, \method retains visible positive MaxSim contributions
within both marked evidence regions, whereas SAP retains a contribution
only within the lower region. As the budget decreases, \method continues
to preserve a visible contribution within the upper evidence region at
$\gamma=0.10$ and within the lower evidence region at $\gamma=0.05$.
In contrast, SAP exhibits no visible contribution within either marked
region under these two more aggressive settings. Although the dominant
response shifts across compression budgets, the consistently higher
Pair-SR and continued overlap with query-relevant evidence are consistent
with the intended effect of folding: preserving matching signals that
may otherwise be removed by hard pruning.

\section{Conclusion}

In this work, we introduce \method, a training-free framework for document-side index compression in multi-vector visual document retrieval. \method organizes compression into a focus-then-fold process: Recursive Attention Propagation identifies high-centrality anchors, while similarity-based assignment and centrality-weighted aggregation consolidate the remaining visual evidence into compact representations. Across ViDoRe v1/v2 and REAL-MM-RAG with three heterogeneous backbones, \method consistently outperforms all evaluated training-free baselines from $5\times$ to $20\times$ compression, with larger margins under stronger compression. Ultimately, \method advances the practical scalability of multi-vector VDR by enabling high-fidelity index compression without additional training.

\bibliography{aaai2027}

@inproceedings{
  faysse2024colpali,
  title={ColPali: Efficient Document Retrieval with Vision Language Models},
  author={Manuel Faysse and Hugues Sibille and Tony Wu and Bilel Omrani and Gautier Viaud and CELINE HUDELOT and Pierre Colombo},
  booktitle={The Thirteenth International Conference on Learning Representations},
  year={2025},
  url={https://openreview.net/forum?id=ogjBpZ8uSi}
}

@inproceedings{khattab2020colbert,
  author = {Khattab, Omar and Zaharia, Matei},
  title = {ColBERT: Efficient and Effective Passage Search via Contextualized Late Interaction over BERT},
  year = {2020},
  isbn = {9781450380164},
  publisher = {Association for Computing Machinery},
  address = {New York, NY, USA},
  url = {https://doi.org/10.1145/3397271.3401075},
  doi = {10.1145/3397271.3401075},
  booktitle = {Proceedings of the 43rd International ACM SIGIR Conference on Research and Development in Information Retrieval},
  pages = {39–48},
  numpages = {10},
  location = {Virtual Event, China},
  series = {SIGIR '20}
}

@misc{mace2025vidorev2,
  title={ViDoRe Benchmark V2: Raising the Bar for Visual Retrieval}, 
  author={Quentin Macé and António Loison and Manuel Faysse},
  year={2025},
  eprint={2505.17166},
  archivePrefix={arXiv},
  primaryClass={cs.IR},
  url={https://arxiv.org/abs/2505.17166}, 
}

@inproceedings{wasserman2025realmmrag,
  title={Real-mm-rag: A real-world multi-modal retrieval benchmark},
  author={Wasserman, Navve and Pony, Roi and Naparstek, Oshri and Goldfarb, Adi Raz and Schwartz, Eli and Barzelay, Udi and Karlinsky, Leonid},
  booktitle={Proceedings of the 63rd Annual Meeting of the Association for Computational Linguistics (Volume 1: Long Papers)},
  pages={31660--31683},
  year={2025}
}

@misc{bai2025qwen25vl,
  title={Qwen2.5-VL Technical Report},
  author={Shuai Bai and Keqin Chen and Xuejing Liu and Jialin Wang and Wenbin Ge and Sibo Song and Kai Dang and Peng Wang and Shijie Wang and Jun Tang and Humen Zhong and Yuanzhi Zhu and Mingkun Yang and Zhaohai Li and Jianqiang Wan and Pengfei Wang and Wei Ding and Zheren Fu and Yiheng Xu and Jiabo Ye and Xi Zhang and Tianbao Xie and Zesen Cheng and Hang Zhang and Zhibo Yang and Haiyang Xu and Junyang Lin},
  year={2025},
  eprint={2502.13923},
  archivePrefix={arXiv},
  primaryClass={cs.CV},
  url={https://arxiv.org/abs/2502.13923},
}

@misc{moreira2026nemotron,
  title={Nemotron ColEmbed V2: Top-Performing Late Interaction Embedding Models for Visual Document Retrieval}, 
  author={Gabriel de Souza P. Moreira and Ronay Ak and Mengyao Xu and Oliver Holworthy and Benedikt Schifferer and Zhiding Yu and Yauhen Babakhin and Radek Osmulski and Jiarui Cai and Ryan Chesler and Bo Liu and Even Oldridge},
  year={2026},
  eprint={2602.03992},
  archivePrefix={arXiv},
  primaryClass={cs.IR},
  url={https://arxiv.org/abs/2602.03992}, 
}

@misc{ma2025lightcolpali,
  title={Towards Storage-Efficient Visual Document Retrieval: An Empirical Study on Reducing Patch-Level Embeddings}, 
  author={Yubo Ma and Jinsong Li and Yuhang Zang and Xiaobao Wu and Xiaoyi Dong and Pan Zhang and Yuhang Cao and Haodong Duan and Jiaqi Wang and Yixin Cao and Aixin Sun},
  year={2025},
  eprint={2506.04997},
  archivePrefix={arXiv},
  primaryClass={cs.IR},
  url={https://arxiv.org/abs/2506.04997}, 
}

@misc{yan2025docpruner,
      title={DocPruner: A Storage-Efficient Framework for Multi-Vector Visual Document Retrieval via Adaptive Patch-Level Embedding Pruning}, 
      author={Yibo Yan and Guangwei Xu and Xin Zou and Shuliang Liu and James Kwok and Xuming Hu},
      year={2025},
      eprint={2509.23883},
      archivePrefix={arXiv},
      primaryClass={cs.CL},
      url={https://arxiv.org/abs/2509.23883}, 
}

@misc{liu2026sap,
  title={Structural Anchor Pruning: Training-Free Multi-Vector Compression for Visual Document Retrieval}, 
  author={Zhuchenyang Liu and Ziyu Hu and Yao Zhang and Yu Xiao},
  year={2026},
  eprint={2601.20107},
  archivePrefix={arXiv},
  primaryClass={cs.CV},
  url={https://arxiv.org/abs/2601.20107}, 
}

@inproceedings{yan2026prunethenmerge,
  title={Sculpting the Vector Space: Towards Efficient Multi-Vector Visual Document Retrieval via Prune-then-Merge Framework},
  author={Yan, Yibo and Ou, Mingdong and Cao, Yi and Zou, Xin and Huo, Jiahao and Liu, Shuliang and Kwok, James and Hu, Xuming},
  booktitle={Findings of the Association for Computational Linguistics: ACL 2026},
  pages={24883--24925},
  year={2026}
}

@misc{yan2026colchunk,
      title={Visual Late Chunking: An Empirical Study of Contextual Chunking for Efficient Visual Document Retrieval}, 
      author={Yibo Yan and Mingdong Ou and Yi Cao and Jiahao Huo and Xin Zou and Shuliang Liu and James Kwok and Xuming Hu},
      year={2026},
      eprint={2604.10167},
      archivePrefix={arXiv},
      primaryClass={cs.CV},
      url={https://arxiv.org/abs/2604.10167}, 
}

@inproceedings{xiao2025metaembed,
  title={MetaEmbed: Scaling Multimodal Retrieval at Test-Time with Flexible Late Interaction},
  author={Zilin Xiao and Qi Ma and Mengting Gu and Chun-cheng Jason Chen and Xintao Chen and Vicente Ordonez and Vijai Mohan},
  booktitle={The Fourteenth International Conference on Learning Representations},
  year={2026},
  url={https://openreview.net/forum?id=yKDqg9HwZX}
}

@misc{huo2026causalembed,
  title={CausalEmbed: Auto-Regressive Multi-Vector Generation in Latent Space for Visual Document Embedding}, 
  author={Jiahao Huo and Yu Huang and Yibo Yan and Ye Pan and Yi Cao and Mingdong Ou and Philip S. Yu and Xuming Hu},
  year={2026},
  eprint={2601.21262},
  archivePrefix={arXiv},
  primaryClass={cs.CL},
  url={https://arxiv.org/abs/2601.21262}, 
}

@misc{qin2026agc,
      title={Multi-Vector Index Compression in Any Modality}, 
      author={Hanxiang Qin and Alexander Martin and Rohan Jha and Chunsheng Zuo and Reno Kriz and Benjamin Van Durme},
      year={2026},
      eprint={2602.21202},
      archivePrefix={arXiv},
      primaryClass={cs.IR},
      url={https://arxiv.org/abs/2602.21202}, 
}

@misc{yan2026unlocking,
      title={Unlocking Multimodal Document Intelligence: From Current Triumphs to Future Frontiers of Visual Document Retrieval}, 
      author={Yibo Yan and Jiahao Huo and Guanbo Feng and Mingdong Ou and Yi Cao and Xin Zou and Shuliang Liu and Yuanhuiyi Lyu and Yu Huang and Jungang Li and Kening Zheng and Xu Zheng and Philip S. Yu and James Kwok and Xuming Hu},
      year={2026},
      eprint={2602.19961},
      archivePrefix={arXiv},
      primaryClass={cs.CL},
      url={https://arxiv.org/abs/2602.19961}, 
}

@misc{xiang2026mm,
      title={MM-Matryoshka: Towards Budget-Elastic Visual Document Retrieval via a 2D Multimodal Matryoshka Training Framework}, 
      author={Haowen Xiang and Yibo Yan and Jiahao Huo and Yu Huang and Yi Cao and Mingdong Ou and Xuming Hu},
      year={2026},
      eprint={2606.07654},
      archivePrefix={arXiv},
      primaryClass={cs.CV},
      url={https://arxiv.org/abs/2606.07654}, 
}

@misc{yan2026colparse,
      title={Beyond the Grid: Layout-Informed Multi-Vector Retrieval with Parsed Visual Document Representations}, 
      author={Yibo Yan and Mingdong Ou and Yi Cao and Xin Zou and Shuliang Liu and Jiahao Huo and Yu Huang and James Kwok and Xuming Hu},
      year={2026},
      eprint={2603.01666},
      archivePrefix={arXiv},
      primaryClass={cs.CL},
      url={https://arxiv.org/abs/2603.01666}, 
}

@inproceedings{wang2023zerotprune,
  title={Zero-tprune: Zero-shot token pruning through leveraging of the attention graph in pre-trained transformers},
  author={Wang, Hongjie and Dedhia, Bhishma and Jha, Niraj K},
  booktitle={Proceedings of the IEEE/CVF Conference on Computer Vision and Pattern Recognition},
  pages={16070--16079},
  year={2024}
}

@inproceedings{ma2024dse,
  title={Unifying multimodal retrieval via document screenshot embedding},
  author={Ma, Xueguang and Lin, Sheng-Chieh and Li, Minghan and Chen, Wenhu and Lin, Jimmy},
  booktitle={Proceedings of the 2024 Conference on Empirical Methods in Natural Language Processing},
  pages={6492--6505},
  year={2024}
}

@misc{gunther2025jinav4,
  title={jina-embeddings-v4: Universal Embeddings for Multimodal Multilingual Retrieval}, 
  author={Michael Günther and Saba Sturua and Mohammad Kalim Akram and Isabelle Mohr and Andrei Ungureanu and Sedigheh Eslami and Scott Martens and Bo Wang and Nan Wang and Han Xiao},
  year={2025},
  eprint={2506.18902},
  archivePrefix={arXiv},
  primaryClass={cs.AI},
  url={https://arxiv.org/abs/2506.18902}, 
}

@inproceedings{acquavia2023staticpruning,
  author = {Acquavia, Antonio and Macdonald, Craig and Tonellotto, Nicola},
  title = {Static Pruning for Multi-Representation Dense Retrieval},
  year = {2023},
  isbn = {9798400700279},
  publisher = {Association for Computing Machinery},
  address = {New York, NY, USA},
  url = {https://doi.org/10.1145/3573128.3604896},
  doi = {10.1145/3573128.3604896},
  booktitle = {Proceedings of the ACM Symposium on Document Engineering 2023},
  articleno = {7},
  numpages = {10},
  location = {Limerick, Ireland},
  series = {DocEng '23}
}

@inproceedings{lassance2022learnedtokenpruning,
  title={Learned token pruning in contextualized late interaction over BERT (ColBERT)},
  author={Lassance, Carlos and Maachou, Maroua and Park, Joohee and Clinchant, St{\'e}phane},
  booktitle={Proceedings of the 45th International ACM SIGIR Conference on Research and Development in Information Retrieval},
  pages={2232--2236},
  year={2022}
}

@inproceedings{zong2025losslesspruning,
  title={Towards Lossless Token Pruning in Late-Interaction Retrieval Models},
  author={Zong, Yuxuan and Piwowarski, Benjamin},
  booktitle={Proceedings of the 48th International ACM SIGIR Conference on Research and Development in Information Retrieval},
  pages={2407--2417},
  year={2025}
}

@misc{clavie2024tokenpooling,
      title={Reducing the Footprint of Multi-Vector Retrieval with Minimal Performance Impact via Token Pooling}, 
      author={Benjamin Clavié and Antoine Chaffin and Griffin Adams},
      year={2024},
      eprint={2409.14683},
      archivePrefix={arXiv},
      primaryClass={cs.IR},
      url={https://arxiv.org/abs/2409.14683}, 
}

@inproceedings{macavaney2025constantspace,
  title={Efficient constant-space multi-vector retrieval},
  author={MacAvaney, Sean and Mallia, Antonio and Tonellotto, Nicola},
  booktitle={European Conference on Information Retrieval},
  pages={237--245},
  year={2025},
  organization={Springer}
}

@inproceedings{wang2013ndcg,
  title={A theoretical analysis of NDCG type ranking measures},
  author={Wang, Yining and Wang, Liwei and Li, Yuanzhi and He, Di and Liu, Tie-Yan},
  booktitle={Conference on learning theory},
  pages={25--54},
  year={2013},
  organization={PMLR}
}

@inproceedings{santhanam2022colbertv2,
  title={Colbertv2: Effective and efficient retrieval via lightweight late interaction},
  author={Santhanam, Keshav and Khattab, Omar and Saad-Falcon, Jon and Potts, Christopher and Zaharia, Matei},
  booktitle={Proceedings of the 2022 Conference of the North American Chapter of the Association for Computational Linguistics: Human Language Technologies},
  pages={3715--3734},
  year={2022}
}

@inproceedings{nardini2024emvb,
  title={Efficient Multi-vector Dense Retrieval with Bit Vectors},
  author={Nardini, Franco Maria and Rulli, Cosimo and Venturini, Rossano},
  booktitle={European Conference on Information Retrieval},
  pages={3--17},
  year={2024},
  organization={Springer}
}

@misc{veneroso2025crisp,
      title={CRISP: Clustering Multi-Vector Representations for Denoising and Pruning}, 
      author={João Veneroso and Rajesh Jayaram and Jinmeng Rao and Gustavo Hernández Ábrego and Majid Hadian and Daniel Cer},
      year={2025},
      eprint={2505.11471},
      archivePrefix={arXiv},
      primaryClass={cs.IR},
      url={https://arxiv.org/abs/2505.11471}, 
}

@inproceedings{wang2025vidorag,
  title={Vidorag: Visual document retrieval-augmented generation via dynamic iterative reasoning agents},
  author={Wang, Qiuchen and Ding, Ruixue and Chen, Zehui and Wu, Weiqi and Wang, Shihang and Xie, Pengjun and Zhao, Feng},
  booktitle={Proceedings of the 2025 Conference on Empirical Methods in Natural Language Processing},
  pages={9124--9145},
  year={2025}
}

@inproceedings{tanaka2025vdocrag,
  title={Vdocrag: Retrieval-augmented generation over visually-rich documents},
  author={Tanaka, Ryota and Iki, Taichi and Hasegawa, Taku and Nishida, Kyosuke and Saito, Kuniko and Suzuki, Jun},
  booktitle={Proceedings of the Computer Vision and Pattern Recognition Conference},
  pages={24827--24837},
  year={2025}
}

@misc{jha2026brief,
      title={A Brief Comparison of Training-Free Multi-Vector Sequence Compression Methods}, 
      author={Rohan Jha and Chunsheng Zuo and Reno Kriz and Benjamin Van Durme},
      year={2026},
      eprint={2603.22434},
      archivePrefix={arXiv},
      primaryClass={cs.IR},
      url={https://arxiv.org/abs/2603.22434}, 
}

@inproceedings{archish2026incorporating,
  title={Incorporating token importance in multi-vector retrieval},
  author={Archish, S and Garg, Ankit and Shiragur, Kirankumar and Kayal, Neeraj},
  booktitle={Proceedings of the AAAI Conference on Artificial Intelligence},
  pages={32860--32866},
  year={2026}
}

@misc{jaasaari2026lemur,
      title={LEMUR: Learned Multi-Vector Retrieval}, 
      author={Elias Jääsaari and Ville Hyvönen and Teemu Roos},
      year={2026},
      eprint={2601.21853},
      archivePrefix={arXiv},
      primaryClass={cs.IR},
      url={https://arxiv.org/abs/2601.21853}, 
}

@misc{cha2026reinpool,
      title={ReinPool: Reinforcement Learning Pooling Multi-Vector Embeddings for Retrieval System}, 
      author={Sungguk Cha and DongWook Kim and Mintae Kim and Youngsub Han and Byoung-Ki Jeon and Sangyeob Lee},
      year={2026},
      eprint={2601.07125},
      archivePrefix={arXiv},
      primaryClass={cs.IR},
      url={https://arxiv.org/abs/2601.07125}, 
}

@misc{chaffin2026colbert,
      title={ColBERT-Zero: To Pre-train Or Not To Pre-train ColBERT models}, 
      author={Antoine Chaffin and Luca Arnaboldi and Amélie Chatelain and Florent Krzakala},
      year={2026},
      eprint={2602.16609},
      archivePrefix={arXiv},
      primaryClass={cs.CL},
      url={https://arxiv.org/abs/2602.16609}, 
}

\clearpage
\begin{center}
{\Large\bfseries Technical Appendix}
\end{center}
\appendix

\section*{Contents}

\begingroup
\normalsize
\setlength{\parskip}{1pt}
\setlength{\parindent}{0pt}

\textbf{A\quad Experimental Details}
\hfill \pageref{app:experimental_setup}\par
\hspace*{1em}A.1\quad Benchmarks and Evaluation
\hfill \pageref{app:benchmark_evaluation_details}\par
\hspace*{1em}A.2\quad Retrieval Backbones
\hfill \pageref{app:retrieval_backbone_details}\par
\hspace*{1em}A.3\quad Baselines and Budget Matching
\hfill \pageref{app:baseline_details}\par
\hspace*{1em}A.4\quad \method Configuration
\hfill \pageref{app:anchorfold_configuration}\par
\hspace*{1em}A.5\quad Implementation Details
\hfill \pageref{app:implementation_details}\par

\medskip

\textbf{B\quad Complete Experimental Results}
\hfill \pageref{app:complete_experimental_results}\par
\hspace*{1em}B.1\quad Benchmark-Level Results
\hfill \pageref{app:benchmark_level_complete_results}\par
\hspace*{1em}B.2\quad Detailed Results on ViDoRe v1
\hfill \pageref{app:vidore_v1_complete_results}\par
\hspace*{1em}B.3\quad Detailed Results on ViDoRe v2
\hfill \pageref{app:vidore_v2_complete_results}\par
\hspace*{1em}B.4\quad Detailed Results on REAL-MM-RAG
\hfill \pageref{app:real_mm_rag_complete_results}\par

\medskip

\textbf{C\quad Detailed Comparison with Trained Compression}
\hfill \pageref{app:trained_compression_comparison}\par
\hspace*{1em}C.1\quad Controlled Comparison Protocol
\hfill \pageref{app:trained_compression_protocol}\par
\hspace*{1em}C.2\quad Complete Comparison Results
\hfill \pageref{app:trained_compression_complete_results}\par

\medskip

\textbf{D\quad Calibration Details and Validation}
\hfill \pageref{app:calibration_details}\par
\hspace*{1em}D.1\quad Joint SR-Based Calibration Protocol
\hfill \pageref{app:joint_sr_calibration_protocol}\par
\hspace*{1em}D.2\quad Joint Calibration Results
\hfill \pageref{app:joint_sr_calibration_results}\par
\hspace*{1em}D.3\quad SR-Guided Layer-Window Validation
\hfill \pageref{app:layer_window_selection_validation}\par
\hspace*{1em}D.4\quad Propagation-Depth Validation
\hfill \pageref{app:propagation_depth_selection_validation}\par

\medskip

\textbf{E\quad Efficiency Evaluation Details}
\hfill \pageref{app:efficiency_details}\par
\hspace*{1em}E.1\quad Efficiency Measurement Protocol
\hfill \pageref{app:efficiency_measurement_protocol}\par
\hspace*{1em}E.2\quad Computational Complexity
\hfill \pageref{app:efficiency_complexity}\par
\hspace*{1em}E.3\quad Complete Efficiency Results
\hfill \pageref{app:complete_efficiency_results}\par

\endgroup

\clearpage
\section{Experimental Details}
\label{app:experimental_setup}

\subsection{Benchmarks and Evaluation}
\label{app:benchmark_evaluation_details}

\paragraph{Benchmarks.}
We evaluate on all constituent test sets of the official ViDoRe v1
\citep{faysse2024colpali} and ViDoRe v2
\citep{mace2025vidorev2} benchmarks.
ViDoRe v1 comprises ten datasets spanning scientific figures, scanned
documents, infographics, tables, financial reports, and domain-specific
reports in English and French.

ViDoRe v2 comprises four collections covering ESG reports, biomedical
lectures, and economics reports. Compared with ViDoRe v1, it places
greater emphasis on long-form documents, multi-page relevance, and
queries with reduced lexical overlap with the source pages. Three
collections provide queries in English, French, German, and Spanish,
whereas the fourth contains a fully human-labeled English query set
that is disjoint from its semi-synthetic ESG counterpart. We use the
complete official test split of every constituent dataset without
additional subsampling.

We further evaluate cross-benchmark generalization on the four
BEIR-compatible REAL-MM-RAG collections---FinReport, FinSlides,
TechReport, and TechSlides \citep{wasserman2025realmmrag}. These
collections cover reports and presentation slides in the finance and
technology domains. REAL-MM-RAG provides multiple rephrased query
variants with progressively reduced lexical overlap with the relevant
page. We use the \texttt{rephrase\_level\_3} queries throughout all
REAL-MM-RAG experiments.

\paragraph{Evaluation protocol.}
We use NDCG@5 as the primary retrieval metric. For each constituent
dataset, NDCG@5 is averaged over all test queries. For each benchmark
suite, we report the unweighted macro-average over its constituent
datasets.

To quantify retrieval fidelity relative to the corresponding
uncompressed index, we additionally report NDCG@5 retention:
\begin{equation}
    \operatorname{Retention}_{b,\mathcal{B},c}
    =
    100
    \times
    \frac{
        \mathrm{NDCG@5}_{b,\mathcal{B},c}
    }{
        \mathrm{NDCG@5}_{b,\mathcal{B},\mathrm{Full}}
    },
\end{equation}
where \(b\) denotes the retrieval backbone, \(\mathcal{B}\) the
benchmark suite, and \(c\) the compression method. When aggregating
retention across multiple backbone--benchmark pairs, we first compute
retention independently for each pair and then take their unweighted
arithmetic mean.

\subsection{Retrieval Backbones}
\label{app:retrieval_backbone_details}

We evaluate three late-interaction retrievers that differ substantially
in backbone architecture, depth, visual tokenization, and retrieval
embedding dimensionality, as summarized in
Table~\ref{tab:retrieval_backbones}.

\paragraph{ColPali v1.3.}
We use the official \texttt{vidore/colpali-v1.3} checkpoint
\citep{faysse2024colpali} without further training.

\paragraph{ColQwen2.5.}
We construct a bidirectional ColQwen2.5 retriever from
\texttt{vidore/colqwen2.5-base} \citep{bai2025qwen25vl}.
Specifically, we remove the causal mask from self-attention throughout
the language backbone and train the resulting retriever on the official
\texttt{vidore/colpali\_train\_set}, following the training recipe
released with \texttt{vidore/colqwen2.5-v0.2}. We use the official
three-epoch LoRA configuration with a per-device batch size of 32, a
learning rate of \(10^{-4}\), 500 warm-up steps,
\(r=\alpha=32\), a LoRA dropout of \(0.1\), bfloat16 precision, and
FlashAttention-2.

\paragraph{Nemotron ColEmbed-3B v2.}
We use the official
\texttt{nvidia/llama-nemotron-colembed-vl-3b-v2}
checkpoint \citep{moreira2026nemotron} without further training.

\begin{table*}[t]
    \centering
    \small
    \setlength{\tabcolsep}{3.5pt}
    
    \begin{tabular}{@{}lcccl@{}}
        \toprule
        Retriever
        & Language backbone
        & Layers
        & Retrieval dim.
        & Image processing \\
        \midrule
        ColPali v1.3
        & PaliGemma-3B
        & 18
        & 128
        & Fixed \(448\times448\); 1,024 visual tokens \\
        ColQwen2.5
        & Qwen2.5-VL-3B
        & 36
        & 128
        & Dynamic resolution; \(\leq768\) visual tokens \\
        Nemotron ColEmbed-3B v2
        & Llama-3.2-3B
        & 28
        & 3,072
        & \(512\times512\) dynamic tiling; \(\leq8\) tiles + thumbnail \\
        \bottomrule
    \end{tabular}

    \caption{
    Retrieval backbones used in our experiments. The layer count refers
    to the language backbone.
    }
    \label{tab:retrieval_backbones}
\end{table*}

\subsection{Baselines and Budget Matching}
\label{app:baseline_details}

All baselines are reproduced within the same evaluation pipeline and
operate exclusively on document-side representations. Query encoding
and the original MaxSim scoring function remain unchanged.

\paragraph{Full Index.}
The uncompressed reference retains all valid visual retrieval vectors
produced by the corresponding backbone.

\paragraph{Structural Anchor Pruning.}
SAP \citep{liu2026sap} restricts intermediate-layer self-attention to
valid visual tokens and ranks them using visual in-degree centrality
aggregated across attention heads and a selected layer window. Given a
budget \(m\), SAP retains the \(m\) highest-scoring visual vectors
without modifying their representations. We independently apply SAP's
label-free SR-guided window-selection procedure to each retrieval
backbone and use the resulting calibrated window for all evaluation
datasets and retention ratios. The selected zero-indexed windows are
11--14 for ColPali, 26--33 for ColQwen2.5, and 18--23 for Nemotron
ColEmbed-3B v2.

\paragraph{DocPruner.}
DocPruner \citep{yan2025docpruner} uses final-layer global-token
attention to adapt the number of retained visual vectors to each page.
To match a prescribed global retention ratio, we standardize the
attention scores within each calibration page and, for each target
\(\gamma\), determine a single global threshold from the corresponding
empirical quantile over 128 held-out calibration pages. This preserves
DocPruner's document-adaptive allocation while matching its global
average retention rate to the target budget. The calibration pages are
disjoint from all evaluation splits.

\paragraph{Semantic Clustering.}
Following \citet{ma2025lightcolpali}, we apply \(K\)-means independently
to the post-projection visual retrieval vectors of each page and
represent each cluster by the arithmetic mean of its members. The
number of clusters is set directly by the common per-page output
budget.

\paragraph{1D-Pooling.}
Following \citet{ma2025lightcolpali}, we partition the ordered sequence
of visual retrieval vectors into contiguous, non-overlapping groups
and represent each group by the arithmetic mean of its members. Group
boundaries are chosen such that the number of output vectors matches
the common per-page budget.

\paragraph{Budget matching.}
We evaluate retention ratios
\(\gamma\in\{0.50,0.20,0.10,0.05\}\), corresponding to nominal
document-vector compression factors of \(2\times\), \(5\times\),
\(10\times\), and \(20\times\), respectively. For \method and all
fixed-budget baselines, a page containing \(N\) valid visual vectors is
represented using the common budget
\begin{equation}
    m=\left\lceil\gamma N\right\rceil.
\end{equation}
For Semantic Clustering, \(m\) specifies the number of clusters; for
1D-Pooling, it specifies the number of contiguous groups. DocPruner is
matched to the same target global retention ratio using the held-out
quantile calibration described above.

\subsection{\method Configuration}
\label{app:anchorfold_configuration}

The configurations used in the reported experiments are selected
independently for each retrieval backbone using the label-free joint
SR-based calibration procedure described in
Appendix~\ref{app:calibration_details}. The calibration selects
propagation depth \(K=6\) for all three backbones and yields the
zero-indexed layer windows 11--14 for ColPali v1.3, 25--32 for
ColQwen2.5, and 14--19 for Nemotron ColEmbed-3B v2. These calibrated
configurations are then fixed across all benchmarks and retention
ratios.

\subsection{Implementation Details}
\label{app:implementation_details}

All methods are implemented within a unified PyTorch evaluation
pipeline. We use the official model-specific image processors and
preserve each backbone's native image-processing and visual-tokenization
policy. No OCR text or externally provided layout annotations are used
by any compression method. Evaluation queries are not available to the
document compressor during index construction; queries are used only
for the one-time, label-free calibration on the held-out query--page
pairs described in Appendix~\ref{app:joint_sr_calibration_protocol}.

After backbone construction or retrieval training, all retriever
parameters remain frozen during compression and evaluation. Compression
is applied once during offline document indexing, and the compressed
representations are reused for all subsequent queries with the original
query encoder and MaxSim scoring function. Backbone inference uses
bfloat16 precision. All experiments are conducted on a cluster equipped
with NVIDIA A800 80-GB GPUs.

\section{Complete Experimental Results}
\label{app:complete_experimental_results}

This section reports the complete benchmark-level and per-dataset results
underlying the comparisons in the main paper. All NDCG@5 values are reported
on a 0--100 scale. For each backbone--dataset pair, NDCG@5 retention is
measured relative to the corresponding Full Index, following
Appendix~\ref{app:benchmark_evaluation_details}. At the benchmark level, we first compute the unweighted macro-average
NDCG@5 over the constituent datasets and then calculate retention with respect
to the corresponding Full Index macro-average. Accordingly, benchmark-level
retention is not obtained by averaging the per-dataset retention values.

\subsection{Benchmark-Level Results}
\label{app:benchmark_level_complete_results}

Table~\ref{tab:complete_benchmark_results} summarizes the complete
benchmark-level results for all three retrieval backbones, the five evaluated
training-free compression methods, and four matched document-vector retention
ratios. For each backbone--benchmark pair, Full gives the uncompressed
reference NDCG@5, while every compressed setting reports both NDCG@5 and its
retention relative to Full.

\begin{table*}[t]
    \centering
    \small
    \setlength{\tabcolsep}{1.5pt}
    \renewcommand{\arraystretch}{0.98}
    
    \begin{tabular*}{\textwidth}{@{\extracolsep{\fill}}cc*{9}{c}@{}}
        \toprule
        Backbone & Method & Full & \multicolumn{2}{c}{$\gamma=0.50$} & \multicolumn{2}{c}{$\gamma=0.20$} & \multicolumn{2}{c}{$\gamma=0.10$} & \multicolumn{2}{c}{$\gamma=0.05$} \\
        \cmidrule(lr){3-3}\cmidrule(lr){4-5}\cmidrule(lr){6-7}\cmidrule(lr){8-9}\cmidrule(lr){10-11}
        & & NDCG@5 & NDCG@5 & Ret. (\%) & NDCG@5 & Ret. (\%) & NDCG@5 & Ret. (\%) & NDCG@5 & Ret. (\%) \\
        \midrule
        \multicolumn{11}{@{}l}{\textbf{ViDoRe v1} \textit{(10 datasets)}} \\
        \addlinespace[1pt]
        \multirow{5}{*}{ColPali} & 1D-Pooling & \multirow{5}{*}{83.7} & 82.1 & 98.1 & 77.9 & 93.1 & 72.7 & 86.9 & 68.3 & 81.6 \\
         & Semantic Clust. &  & \textbf{83.2} & \textbf{99.4} & 81.2 & 97.0 & 79.4 & 94.9 & 76.6 & 91.5 \\
         & DocPruner &  & 79.4 & 94.9 & 76.2 & 91.0 & 69.5 & 83.0 & 63.4 & 75.7 \\
         & SAP &  & 82.8 & 98.9 & 81.4 & 97.3 & 78.0 & 93.2 & 72.3 & 86.4 \\
         & \textbf{\method} &  & \textbf{83.2} & \textbf{99.4} & \textbf{81.9} & \textbf{97.8} & \textbf{79.9} & \textbf{95.5} & \textbf{76.7} & \textbf{91.6} \\
        \cmidrule(lr){1-11}
        \multirow{5}{*}{ColQwen2.5} & 1D-Pooling & \multirow{5}{*}{89.6} & 87.5 & 97.7 & 82.3 & 91.9 & 78.2 & 87.3 & 74.6 & 83.3 \\
         & Semantic Clust. &  & 89.0 & 99.3 & 87.5 & 97.7 & 86.0 & 96.0 & 82.6 & 92.2 \\
         & DocPruner &  & 87.6 & 97.8 & 78.8 & 87.9 & 73.4 & 81.9 & 66.6 & 74.3 \\
         & SAP &  & \textbf{89.4} & \textbf{99.8} & 87.2 & 97.3 & 84.4 & 94.2 & 80.2 & 89.5 \\
         & \textbf{\method} &  & 89.3 & 99.7 & \textbf{88.5} & \textbf{98.8} & \textbf{86.6} & \textbf{96.7} & \textbf{83.7} & \textbf{93.4} \\
        \cmidrule(lr){1-11}
        \multirow{5}{*}{Nemotron ColEmbed-3B} & 1D-Pooling & \multirow{5}{*}{91.7} & 90.5 & 98.7 & 89.3 & 97.4 & 86.8 & 94.7 & 84.7 & 92.4 \\
         & Semantic Clust. &  & 90.7 & 98.9 & 90.1 & 98.3 & 89.3 & 97.4 & 87.3 & 95.2 \\
         & DocPruner &  & 90.5 & 98.7 & 88.7 & 96.7 & 86.4 & 94.2 & 81.9 & 89.3 \\
         & SAP &  & 90.9 & 99.1 & 90.4 & 98.6 & 89.3 & 97.4 & 86.7 & 94.5 \\
         & \textbf{\method} &  & \textbf{91.0} & \textbf{99.2} & \textbf{90.7} & \textbf{98.9} & \textbf{90.3} & \textbf{98.5} & \textbf{88.6} & \textbf{96.6} \\
        \midrule
        \multicolumn{11}{@{}l}{\textbf{ViDoRe v2} \textit{(4 datasets)}} \\
        \addlinespace[1pt]
        \multirow{5}{*}{ColPali} & 1D-Pooling & \multirow{5}{*}{54.4} & 49.0 & 90.1 & 42.9 & 78.9 & 35.7 & 65.6 & 32.0 & 58.8 \\
         & Semantic Clust. &  & 53.2 & 97.8 & 49.2 & 90.4 & 44.8 & 82.4 & 40.7 & 74.8 \\
         & DocPruner &  & 49.0 & 90.1 & 47.0 & 86.4 & 41.4 & 76.1 & 33.8 & 62.1 \\
         & SAP &  & 53.7 & 98.7 & 48.6 & 89.3 & 47.8 & 87.9 & 42.6 & 78.3 \\
         & \textbf{\method} &  & \textbf{53.9} & \textbf{99.1} & \textbf{52.9} & \textbf{97.2} & \textbf{50.2} & \textbf{92.3} & \textbf{47.3} & \textbf{86.9} \\
        \cmidrule(lr){1-11}
        \multirow{5}{*}{ColQwen2.5} & 1D-Pooling & \multirow{5}{*}{61.3} & 54.9 & 89.6 & 43.6 & 71.1 & 37.6 & 61.3 & 33.0 & 53.8 \\
         & Semantic Clust. &  & 58.4 & 95.3 & 53.1 & 86.6 & 48.0 & 78.3 & 41.9 & 68.4 \\
         & DocPruner &  & 60.7 & 99.0 & 52.1 & 85.0 & 48.3 & 78.8 & 35.8 & 58.4 \\
         & SAP &  & 59.8 & 97.6 & 57.1 & 93.1 & 53.7 & 87.6 & 50.1 & 81.7 \\
         & \textbf{\method} &  & \textbf{61.1} & \textbf{99.7} & \textbf{60.3} & \textbf{98.4} & \textbf{57.5} & \textbf{93.8} & \textbf{54.7} & \textbf{89.2} \\
        \cmidrule(lr){1-11}
        \multirow{5}{*}{Nemotron ColEmbed-3B} & 1D-Pooling & \multirow{5}{*}{63.4} & 60.8 & 95.9 & 55.5 & 87.5 & 50.8 & 80.1 & 46.8 & 73.8 \\
         & Semantic Clust. &  & 61.5 & 97.0 & 59.4 & 93.7 & 56.7 & 89.4 & 52.5 & 82.8 \\
         & DocPruner &  & 62.1 & 97.9 & 61.2 & 96.5 & 58.9 & 92.9 & 54.3 & 85.6 \\
         & SAP &  & 61.9 & 97.6 & 60.0 & 94.6 & 59.3 & 93.5 & 55.7 & 87.9 \\
         & \textbf{\method} &  & \textbf{63.0} & \textbf{99.4} & \textbf{62.5} & \textbf{98.6} & \textbf{62.3} & \textbf{98.3} & \textbf{61.2} & \textbf{96.5} \\
        \midrule
        \multicolumn{11}{@{}l}{\textbf{REAL-MM-RAG} \textit{(4 datasets)}} \\
        \addlinespace[1pt]
        \multirow{5}{*}{ColPali} & 1D-Pooling & \multirow{5}{*}{53.6} & 47.7 & 89.0 & 39.7 & 74.1 & 32.8 & 61.2 & 27.6 & 51.5 \\
         & Semantic Clust. &  & 52.6 & 98.1 & 48.9 & 91.2 & 44.9 & 83.8 & 39.5 & 73.7 \\
         & DocPruner &  & 43.9 & 81.9 & 40.9 & 76.3 & 33.4 & 62.3 & 27.2 & 50.7 \\
         & SAP &  & 52.5 & 97.9 & 49.0 & 91.4 & 44.4 & 82.8 & 37.7 & 70.3 \\
         & \textbf{\method} &  & \textbf{53.4} & \textbf{99.6} & \textbf{52.0} & \textbf{97.0} & \textbf{49.0} & \textbf{91.4} & \textbf{44.3} & \textbf{82.6} \\
        \cmidrule(lr){1-11}
        \multirow{5}{*}{ColQwen2.5} & 1D-Pooling & \multirow{5}{*}{64.0} & 58.2 & 90.9 & 46.6 & 72.8 & 40.2 & 62.8 & 33.8 & 52.8 \\
         & Semantic Clust. &  & 62.6 & 97.8 & 58.5 & 91.4 & 53.1 & 83.0 & 44.6 & 69.7 \\
         & DocPruner &  & 57.4 & 89.7 & 40.9 & 63.9 & 35.9 & 56.1 & 29.4 & 45.9 \\
         & SAP &  & 62.7 & 98.0 & 57.2 & 89.4 & 50.3 & 78.6 & 41.7 & 65.2 \\
         & \textbf{\method} &  & \textbf{63.4} & \textbf{99.1} & \textbf{61.3} & \textbf{95.8} & \textbf{57.3} & \textbf{89.5} & \textbf{51.1} & \textbf{79.8} \\
        \cmidrule(lr){1-11}
        \multirow{5}{*}{Nemotron ColEmbed-3B} & 1D-Pooling & \multirow{5}{*}{73.8} & 71.9 & 97.4 & 66.5 & 90.1 & 58.8 & 79.7 & 52.5 & 71.1 \\
         & Semantic Clust. &  & 72.8 & 98.6 & 70.8 & 95.9 & 66.8 & 90.5 & 61.0 & 82.7 \\
         & DocPruner &  & 73.0 & 98.9 & 68.9 & 93.4 & 62.8 & 85.1 & 53.9 & 73.0 \\
         & SAP &  & 73.4 & 99.5 & 71.7 & 97.2 & 69.6 & 94.3 & 65.2 & 88.3 \\
         & \textbf{\method} &  & \textbf{73.8} & \textbf{100.0} & \textbf{73.3} & \textbf{99.3} & \textbf{72.2} & \textbf{97.8} & \textbf{70.5} & \textbf{95.5} \\
        \bottomrule
    \end{tabular*}

    \caption{Complete benchmark-level results. Full is the uncompressed NDCG@5. For each document-vector retention ratio $\gamma$, we report compressed-index NDCG@5 and NDCG@5 retention (\%) relative to Full. Benchmark scores are unweighted macro-averages over constituent datasets.}
    \label{tab:complete_benchmark_results}
\end{table*}

\subsection{Detailed Results on ViDoRe v1}
\label{app:vidore_v1_complete_results}

Tables~\ref{tab:vidore_v1_colpali_complete},
\ref{tab:vidore_v1_colqwen25_complete}, and
\ref{tab:vidore_v1_nemotron_complete} report the ViDoRe v1 per-dataset results
for ColPali v1.3, ColQwen2.5, and Nemotron ColEmbed-3B v2, respectively. In
each table, the Full Index row provides the dataset-specific reference NDCG@5
used to compute retention.

\begin{table*}[t]
    \centering
    \small
    \setlength{\tabcolsep}{2.4pt}
    \renewcommand{\arraystretch}{1.04}
    \begin{tabular*}{\textwidth}{@{\extracolsep{\fill}}cc*{5}{cc}@{}}
        \toprule
        Method & $\gamma$ & \multicolumn{2}{c}{\shortstack{Arxiv\\QA}} & \multicolumn{2}{c}{\shortstack{Doc\\VQA}} & \multicolumn{2}{c}{\shortstack{Info\\VQA}} & \multicolumn{2}{c}{\shortstack{TabF\\Quad}} & \multicolumn{2}{c}{\shortstack{TAT-\\DQA}} \\
        \cmidrule(lr){3-4}\cmidrule(lr){5-6}\cmidrule(lr){7-8}\cmidrule(lr){9-10}\cmidrule(lr){11-12}
        & & NDCG & Ret. & NDCG & Ret. & NDCG & Ret. & NDCG & Ret. & NDCG & Ret. \\
        \midrule
        Full Index & -- & 82.5 & 100.0 & 53.9 & 100.0 & 84.8 & 100.0 & 86.4 & 100.0 & 69.5 & 100.0 \\
        \midrule
        \multirow{4}{*}{1D-Pooling} & 0.50 & 80.7 & 97.8 & 51.1 & 94.9 & 83.6 & 98.5 & 86.7 & 100.3 & 64.6 & 93.0 \\
         & 0.20 & 76.0 & 92.1 & 43.4 & 80.5 & 79.8 & 94.1 & 85.1 & 98.4 & 58.8 & 84.6 \\
         & 0.10 & 71.7 & 86.9 & 35.7 & 66.2 & 77.6 & 91.5 & 80.9 & 93.6 & 53.7 & 77.3 \\
         & 0.05 & 68.1 & 82.5 & 31.0 & 57.4 & 74.5 & 87.9 & 78.9 & 91.3 & 47.9 & 69.0 \\
        \cmidrule(lr){1-12}
        \multirow{4}{*}{Semantic Clust.} & 0.50 & 81.8 & 99.2 & 53.3 & 98.8 & 84.5 & 99.7 & 86.1 & 99.7 & 68.1 & 98.0 \\
         & 0.20 & 81.3 & 98.5 & 51.6 & 95.6 & 82.9 & 97.8 & 86.4 & 100.0 & 64.1 & 92.3 \\
         & 0.10 & 80.8 & 98.0 & 48.6 & 90.1 & 82.0 & 96.7 & 85.0 & 98.3 & 61.2 & 88.1 \\
         & 0.05 & 79.7 & 96.5 & 44.0 & 81.6 & 79.1 & 93.3 & 84.1 & 97.3 & 54.9 & 79.0 \\
        \cmidrule(lr){1-12}
        \multirow{4}{*}{DocPruner} & 0.50 & 79.9 & 96.9 & 51.0 & 94.6 & 80.1 & 94.5 & 86.1 & 99.6 & 61.1 & 88.0 \\
         & 0.20 & 77.5 & 94.0 & 45.9 & 85.2 & 76.5 & 90.2 & 85.3 & 98.7 & 57.1 & 82.1 \\
         & 0.10 & 74.3 & 90.1 & 38.0 & 70.6 & 71.5 & 84.3 & 80.8 & 93.4 & 49.2 & 70.8 \\
         & 0.05 & 69.4 & 84.1 & 31.8 & 58.9 & 66.7 & 78.7 & 74.9 & 86.6 & 40.0 & 57.5 \\
        \cmidrule(lr){1-12}
        \multirow{4}{*}{SAP} & 0.50 & 82.0 & 99.4 & 52.8 & 97.9 & 84.2 & 99.3 & 86.8 & 100.5 & 68.0 & 97.8 \\
         & 0.20 & 79.6 & 96.5 & 50.5 & 93.7 & 82.8 & 97.6 & 85.9 & 99.4 & 63.8 & 91.8 \\
         & 0.10 & 77.3 & 93.6 & 47.2 & 87.5 & 81.0 & 95.5 & 84.1 & 97.3 & 58.1 & 83.6 \\
         & 0.05 & 74.7 & 90.6 & 41.4 & 76.7 & 73.9 & 87.1 & 81.9 & 94.7 & 48.1 & 69.2 \\
        \cmidrule(lr){1-12}
        \multirow{4}{*}{\textbf{\method}} & 0.50 & 81.9 & 99.3 & 53.0 & 98.4 & 83.9 & 98.9 & 87.1 & 100.7 & 68.5 & 98.5 \\
         & 0.20 & 80.2 & 97.1 & 51.5 & 95.6 & 83.1 & 98.0 & 85.9 & 99.4 & 65.9 & 94.8 \\
         & 0.10 & 81.3 & 98.5 & 50.0 & 92.8 & 81.2 & 95.7 & 85.9 & 99.4 & 60.4 & 86.9 \\
         & 0.05 & 79.6 & 96.5 & 44.4 & 82.4 & 77.9 & 91.9 & 83.6 & 96.8 & 53.0 & 76.3 \\
        \bottomrule
    
    \end{tabular*}

    \caption{Per-dataset results on ViDoRe v1 with ColPali v1.3. Each dataset reports NDCG@5 and retention (\%) relative to its Full Index. Benchmark-level macro-averages are reported in Table~\ref{tab:complete_benchmark_results}.}
    \label{tab:vidore_v1_colpali_complete}

\end{table*}

\begin{table*}[t]
    \ContinuedFloat
    \centering
    \small
    \setlength{\tabcolsep}{2.4pt}
    \renewcommand{\arraystretch}{1.04}
    \begin{tabular*}{\textwidth}{@{\extracolsep{\fill}}cc*{5}{cc}@{}}
        \toprule
        Method & $\gamma$ & \multicolumn{2}{c}{\shortstack{Shift\\Project}} & \multicolumn{2}{c}{\shortstack{Syn.\\AI}} & \multicolumn{2}{c}{\shortstack{Syn.\\Energy}} & \multicolumn{2}{c}{\shortstack{Syn.\\Govt}} & \multicolumn{2}{c}{\shortstack{Syn.\\Health}} \\
        \cmidrule(lr){3-4}\cmidrule(lr){5-6}\cmidrule(lr){7-8}\cmidrule(lr){9-10}\cmidrule(lr){11-12}
        & & NDCG & Ret. & NDCG & Ret. & NDCG & Ret. & NDCG & Ret. & NDCG & Ret. \\
        \midrule
        Full Index & -- & 75.4 & 100.0 & 97.4 & 100.0 & 94.7 & 100.0 & 95.3 & 100.0 & 97.0 & 100.0 \\
        \midrule
        \multirow{4}{*}{1D-Pooling} & 0.50 & 74.3 & 98.5 & 96.5 & 99.1 & 93.6 & 98.9 & 93.6 & 98.2 & 95.8 & 98.8 \\
         & 0.20 & 65.4 & 86.7 & 92.4 & 94.8 & 92.4 & 97.6 & 92.6 & 97.1 & 93.3 & 96.2 \\
         & 0.10 & 49.9 & 66.1 & 90.1 & 92.5 & 87.3 & 92.2 & 88.9 & 93.3 & 90.8 & 93.6 \\
         & 0.05 & 45.1 & 59.8 & 87.8 & 90.1 & 79.8 & 84.3 & 83.4 & 87.5 & 86.3 & 89.0 \\
        \cmidrule(lr){1-12}
        \multirow{4}{*}{Semantic Clust.} & 0.50 & 74.8 & 99.2 & 97.1 & 99.6 & 94.3 & 99.6 & 94.3 & 99.0 & 97.3 & 100.3 \\
         & 0.20 & 70.3 & 93.2 & 94.9 & 97.4 & 92.6 & 97.8 & 92.7 & 97.2 & 95.4 & 98.4 \\
         & 0.10 & 63.5 & 84.2 & 94.2 & 96.7 & 92.0 & 97.2 & 92.3 & 96.8 & 93.8 & 96.8 \\
         & 0.05 & 59.6 & 79.1 & 91.4 & 93.8 & 91.6 & 96.8 & 90.1 & 94.5 & 91.8 & 94.7 \\
        \cmidrule(lr){1-12}
        \multirow{4}{*}{DocPruner} & 0.50 & 61.6 & 81.6 & 95.1 & 97.6 & 91.0 & 96.1 & 92.6 & 97.2 & 95.3 & 98.3 \\
         & 0.20 & 60.7 & 80.5 & 92.7 & 95.1 & 86.7 & 91.6 & 88.5 & 92.9 & 91.4 & 94.2 \\
         & 0.10 & 46.3 & 61.4 & 88.4 & 90.7 & 81.2 & 85.7 & 82.4 & 86.5 & 83.0 & 85.6 \\
         & 0.05 & 38.3 & 50.8 & 78.5 & 80.6 & 75.3 & 79.5 & 78.2 & 82.0 & 81.0 & 83.6 \\
        \cmidrule(lr){1-12}
        \multirow{4}{*}{SAP} & 0.50 & 72.7 & 96.3 & 96.3 & 98.8 & 93.6 & 98.8 & 94.3 & 99.0 & 97.0 & 100.1 \\
         & 0.20 & 75.8 & 100.5 & 94.6 & 97.1 & 92.2 & 97.4 & 93.8 & 98.4 & 95.5 & 98.5 \\
         & 0.10 & 67.0 & 88.9 & 92.9 & 95.4 & 88.7 & 93.6 & 90.7 & 95.2 & 92.6 & 95.5 \\
         & 0.05 & 60.4 & 80.1 & 90.5 & 92.9 & 81.4 & 86.0 & 83.1 & 87.1 & 87.5 & 90.2 \\
        \cmidrule(lr){1-12}
        \multirow{4}{*}{\textbf{\method}} & 0.50 & 74.4 & 98.7 & 96.3 & 98.8 & 94.8 & 100.1 & 95.1 & 99.7 & 96.8 & 99.8 \\
         & 0.20 & 73.1 & 96.9 & 96.8 & 99.3 & 92.2 & 97.4 & 95.2 & 99.9 & 95.3 & 98.3 \\
         & 0.10 & 68.0 & 90.2 & 94.8 & 97.2 & 91.2 & 96.3 & 92.0 & 96.5 & 94.3 & 97.3 \\
         & 0.05 & 64.6 & 85.6 & 93.5 & 95.9 & 88.5 & 93.4 & 89.7 & 94.1 & 92.5 & 95.4 \\
        \bottomrule
    
    \end{tabular*}

    \caption[]{Per-dataset results on ViDoRe v1 with ColPali v1.3 (continued).}
\end{table*}

\begin{table*}[t]
    \centering
    \small
    \setlength{\tabcolsep}{2.4pt}
    \renewcommand{\arraystretch}{1.04}
    \begin{tabular*}{\textwidth}{@{\extracolsep{\fill}}cc*{5}{cc}@{}}
        \toprule
        Method & $\gamma$ & \multicolumn{2}{c}{\shortstack{Arxiv\\QA}} & \multicolumn{2}{c}{\shortstack{Doc\\VQA}} & \multicolumn{2}{c}{\shortstack{Info\\VQA}} & \multicolumn{2}{c}{\shortstack{TabF\\Quad}} & \multicolumn{2}{c}{\shortstack{TAT-\\DQA}} \\
        \cmidrule(lr){3-4}\cmidrule(lr){5-6}\cmidrule(lr){7-8}\cmidrule(lr){9-10}\cmidrule(lr){11-12}
        & & NDCG & Ret. & NDCG & Ret. & NDCG & Ret. & NDCG & Ret. & NDCG & Ret. \\
        \midrule
        Full Index & -- & 89.7 & 100.0 & 56.0 & 100.0 & 91.0 & 100.0 & 93.1 & 100.0 & 81.3 & 100.0 \\
        \midrule
        \multirow{4}{*}{1D-Pooling} & 0.50 & 88.1 & 98.2 & 53.0 & 94.6 & 88.2 & 96.9 & 92.8 & 99.7 & 76.1 & 93.6 \\
         & 0.20 & 83.9 & 93.4 & 48.6 & 86.7 & 84.4 & 92.7 & 90.3 & 97.0 & 68.0 & 83.6 \\
         & 0.10 & 79.4 & 88.5 & 43.0 & 76.7 & 81.6 & 89.6 & 86.9 & 93.3 & 62.6 & 77.0 \\
         & 0.05 & 73.9 & 82.4 & 38.5 & 68.8 & 78.1 & 85.8 & 83.6 & 89.8 & 56.6 & 69.6 \\
        \cmidrule(lr){1-12}
        \multirow{4}{*}{Semantic Clust.} & 0.50 & 90.0 & 100.2 & 54.3 & 97.0 & 90.7 & 99.7 & 92.9 & 99.8 & 80.1 & 98.6 \\
         & 0.20 & 89.4 & 99.7 & 53.4 & 95.4 & 89.2 & 98.0 & 92.7 & 99.6 & 77.6 & 95.5 \\
         & 0.10 & 88.4 & 98.5 & 52.1 & 93.0 & 87.6 & 96.2 & 92.7 & 99.5 & 73.7 & 90.6 \\
         & 0.05 & 87.1 & 97.0 & 48.1 & 85.8 & 85.5 & 94.0 & 90.7 & 97.4 & 69.5 & 85.5 \\
        \cmidrule(lr){1-12}
        \multirow{4}{*}{DocPruner} & 0.50 & 88.0 & 98.1 & 52.7 & 94.1 & 87.2 & 95.8 & 92.6 & 99.4 & 80.3 & 98.9 \\
         & 0.20 & 82.9 & 92.4 & 44.6 & 79.6 & 77.7 & 85.4 & 88.5 & 95.0 & 69.8 & 85.8 \\
         & 0.10 & 77.8 & 86.7 & 37.9 & 67.7 & 73.6 & 80.9 & 85.3 & 91.6 & 62.6 & 77.0 \\
         & 0.05 & 72.1 & 80.3 & 32.5 & 58.0 & 68.3 & 75.0 & 78.6 & 84.4 & 54.3 & 66.8 \\
        \cmidrule(lr){1-12}
        \multirow{4}{*}{SAP} & 0.50 & 89.0 & 99.2 & 56.2 & 100.2 & 90.2 & 99.1 & 93.2 & 100.1 & 81.2 & 99.9 \\
         & 0.20 & 87.7 & 97.7 & 54.1 & 96.6 & 87.9 & 96.6 & 92.6 & 99.4 & 77.3 & 95.1 \\
         & 0.10 & 84.8 & 94.5 & 51.8 & 92.5 & 84.7 & 93.1 & 91.5 & 98.2 & 72.6 & 89.3 \\
         & 0.05 & 81.3 & 90.6 & 49.0 & 87.4 & 81.7 & 89.7 & 90.1 & 96.7 & 62.9 & 77.4 \\
        \cmidrule(lr){1-12}
        \multirow{4}{*}{\textbf{\method}} & 0.50 & 89.3 & 99.5 & 54.6 & 97.5 & 90.5 & 99.5 & 93.2 & 100.1 & 81.2 & 99.9 \\
         & 0.20 & 89.1 & 99.3 & 53.7 & 95.8 & 88.6 & 97.3 & 92.7 & 99.6 & 79.7 & 98.1 \\
         & 0.10 & 87.2 & 97.1 & 52.2 & 93.2 & 86.7 & 95.3 & 91.8 & 98.6 & 77.2 & 95.0 \\
         & 0.05 & 85.0 & 94.7 & 48.9 & 87.3 & 84.0 & 92.3 & 91.0 & 97.7 & 71.9 & 88.5 \\
        \bottomrule
    
    \end{tabular*}

    \caption{Per-dataset results on ViDoRe v1 with ColQwen2.5. Each dataset reports NDCG@5 and retention (\%) relative to its Full Index. Benchmark-level macro-averages are reported in Table~\ref{tab:complete_benchmark_results}.}
    \label{tab:vidore_v1_colqwen25_complete}

\end{table*}

\begin{table*}[t]
    \ContinuedFloat
    \centering
    \small
    \setlength{\tabcolsep}{2.4pt}
    \renewcommand{\arraystretch}{1.04}
    \begin{tabular*}{\textwidth}{@{\extracolsep{\fill}}cc*{5}{cc}@{}}
        \toprule
        Method & $\gamma$ & \multicolumn{2}{c}{\shortstack{Shift\\Project}} & \multicolumn{2}{c}{\shortstack{Syn.\\AI}} & \multicolumn{2}{c}{\shortstack{Syn.\\Energy}} & \multicolumn{2}{c}{\shortstack{Syn.\\Govt}} & \multicolumn{2}{c}{\shortstack{Syn.\\Health}} \\
        \cmidrule(lr){3-4}\cmidrule(lr){5-6}\cmidrule(lr){7-8}\cmidrule(lr){9-10}\cmidrule(lr){11-12}
        & & NDCG & Ret. & NDCG & Ret. & NDCG & Ret. & NDCG & Ret. & NDCG & Ret. \\
        \midrule
        Full Index & -- & 89.5 & 100.0 & 99.3 & 100.0 & 97.4 & 100.0 & 99.3 & 100.0 & 99.1 & 100.0 \\
        \midrule
        \multirow{4}{*}{1D-Pooling} & 0.50 & 85.1 & 95.0 & 98.6 & 99.3 & 96.6 & 99.2 & 98.5 & 99.3 & 97.8 & 98.6 \\
         & 0.20 & 71.2 & 79.5 & 94.1 & 94.8 & 94.3 & 96.8 & 93.1 & 93.8 & 95.5 & 96.3 \\
         & 0.10 & 62.1 & 69.4 & 92.2 & 92.9 & 89.1 & 91.5 & 91.5 & 92.2 & 93.6 & 94.4 \\
         & 0.05 & 60.9 & 68.0 & 86.1 & 86.8 & 88.7 & 91.1 & 87.8 & 88.4 & 91.7 & 92.5 \\
        \cmidrule(lr){1-12}
        \multirow{4}{*}{Semantic Clust.} & 0.50 & 88.5 & 98.9 & 98.9 & 99.6 & 96.7 & 99.2 & 98.9 & 99.6 & 98.5 & 99.4 \\
         & 0.20 & 83.1 & 92.8 & 97.9 & 98.6 & 95.3 & 97.8 & 97.4 & 98.1 & 98.5 & 99.4 \\
         & 0.10 & 80.5 & 89.9 & 96.0 & 96.8 & 95.3 & 97.8 & 96.4 & 97.1 & 97.3 & 98.1 \\
         & 0.05 & 69.7 & 77.8 & 95.1 & 95.8 & 91.3 & 93.7 & 92.3 & 92.9 & 96.5 & 97.4 \\
        \cmidrule(lr){1-12}
        \multirow{4}{*}{DocPruner} & 0.50 & 87.6 & 97.9 & 97.2 & 97.9 & 95.2 & 97.7 & 96.8 & 97.5 & 98.2 & 99.0 \\
         & 0.20 & 70.0 & 78.2 & 89.1 & 89.8 & 85.3 & 87.6 & 84.9 & 85.6 & 94.7 & 95.6 \\
         & 0.10 & 58.9 & 65.8 & 82.4 & 83.0 & 82.7 & 84.9 & 80.1 & 80.7 & 92.6 & 93.4 \\
         & 0.05 & 46.7 & 52.2 & 74.9 & 75.5 & 77.4 & 79.4 & 74.3 & 74.9 & 86.6 & 87.4 \\
        \cmidrule(lr){1-12}
        \multirow{4}{*}{SAP} & 0.50 & 89.4 & 99.9 & 99.6 & 100.4 & 97.1 & 99.6 & 98.9 & 99.6 & 99.1 & 100.0 \\
         & 0.20 & 86.3 & 96.4 & 96.5 & 97.2 & 94.8 & 97.3 & 96.4 & 97.1 & 98.8 & 99.6 \\
         & 0.10 & 83.0 & 92.7 & 94.1 & 94.8 & 89.2 & 91.5 & 94.5 & 95.2 & 98.0 & 98.9 \\
         & 0.05 & 76.0 & 84.9 & 89.8 & 90.4 & 84.7 & 87.0 & 92.0 & 92.7 & 94.5 & 95.4 \\
        \cmidrule(lr){1-12}
        \multirow{4}{*}{\textbf{\method}} & 0.50 & 89.0 & 99.4 & 99.6 & 100.4 & 97.1 & 99.7 & 99.6 & 100.4 & 99.1 & 100.0 \\
         & 0.20 & 89.1 & 99.5 & 98.3 & 99.0 & 95.9 & 98.4 & 98.9 & 99.6 & 98.8 & 99.6 \\
         & 0.10 & 86.2 & 96.3 & 96.5 & 97.2 & 92.3 & 94.8 & 98.4 & 99.1 & 97.2 & 98.0 \\
         & 0.05 & 81.9 & 91.5 & 92.9 & 93.6 & 90.7 & 93.1 & 93.9 & 94.6 & 96.4 & 97.3 \\
        \bottomrule
    
    \end{tabular*}

    \caption[]{Per-dataset results on ViDoRe v1 with ColQwen2.5 (continued).}
\end{table*}

\begin{table*}[t]
    \centering
    \small
    \setlength{\tabcolsep}{2.4pt}
    \renewcommand{\arraystretch}{1.04}
    \begin{tabular*}{\textwidth}{@{\extracolsep{\fill}}cc*{5}{cc}@{}}
        \toprule
        Method & $\gamma$ & \multicolumn{2}{c}{\shortstack{Arxiv\\QA}} & \multicolumn{2}{c}{\shortstack{Doc\\VQA}} & \multicolumn{2}{c}{\shortstack{Info\\VQA}} & \multicolumn{2}{c}{\shortstack{TabF\\Quad}} & \multicolumn{2}{c}{\shortstack{TAT-\\DQA}} \\
        \cmidrule(lr){3-4}\cmidrule(lr){5-6}\cmidrule(lr){7-8}\cmidrule(lr){9-10}\cmidrule(lr){11-12}
        & & NDCG & Ret. & NDCG & Ret. & NDCG & Ret. & NDCG & Ret. & NDCG & Ret. \\
        \midrule
        Full Index & -- & 90.4 & 100.0 & 67.2 & 100.0 & 94.7 & 100.0 & 97.3 & 100.0 & 81.0 & 100.0 \\
        \midrule
        \multirow{4}{*}{1D-Pooling} & 0.50 & 90.4 & 100.0 & 58.8 & 87.5 & 92.7 & 97.8 & 97.4 & 100.1 & 79.6 & 98.3 \\
         & 0.20 & 87.6 & 96.9 & 57.7 & 85.9 & 92.1 & 97.2 & 95.9 & 98.6 & 76.8 & 94.8 \\
         & 0.10 & 83.1 & 91.9 & 53.3 & 79.4 & 90.8 & 95.9 & 95.8 & 98.5 & 70.2 & 86.7 \\
         & 0.05 & 80.8 & 89.4 & 51.2 & 76.1 & 89.5 & 94.6 & 94.7 & 97.4 & 62.9 & 77.6 \\
        \cmidrule(lr){1-12}
        \multirow{4}{*}{Semantic Clust.} & 0.50 & 90.2 & 99.7 & 60.0 & 89.2 & 93.2 & 98.4 & 97.2 & 99.9 & 79.9 & 98.7 \\
         & 0.20 & 89.5 & 99.0 & 58.9 & 87.7 & 92.7 & 97.9 & 96.8 & 99.5 & 78.2 & 96.5 \\
         & 0.10 & 88.6 & 98.0 & 57.6 & 85.7 & 92.0 & 97.2 & 96.5 & 99.2 & 76.3 & 94.1 \\
         & 0.05 & 87.9 & 97.3 & 54.8 & 81.6 & 90.8 & 95.9 & 95.5 & 98.1 & 71.0 & 87.6 \\
        \cmidrule(lr){1-12}
        \multirow{4}{*}{DocPruner} & 0.50 & 90.3 & 99.9 & 60.4 & 89.8 & 93.2 & 98.4 & 97.0 & 99.7 & 78.8 & 97.3 \\
         & 0.20 & 88.6 & 98.0 & 59.9 & 89.2 & 91.4 & 96.5 & 96.8 & 99.4 & 71.1 & 87.8 \\
         & 0.10 & 85.7 & 94.8 & 57.5 & 85.6 & 89.5 & 94.5 & 96.7 & 99.3 & 62.3 & 77.0 \\
         & 0.05 & 80.7 & 89.3 & 53.7 & 79.9 & 86.2 & 91.1 & 94.3 & 96.9 & 50.1 & 61.9 \\
        \cmidrule(lr){1-12}
        \multirow{4}{*}{SAP} & 0.50 & 90.3 & 99.9 & 60.3 & 89.7 & 93.1 & 98.3 & 97.0 & 99.7 & 80.5 & 99.3 \\
         & 0.20 & 90.5 & 100.1 & 58.9 & 87.7 & 92.6 & 97.8 & 97.2 & 99.9 & 79.1 & 97.6 \\
         & 0.10 & 88.2 & 97.5 & 58.2 & 86.6 & 90.7 & 95.8 & 96.5 & 99.2 & 75.9 & 93.6 \\
         & 0.05 & 84.9 & 93.9 & 56.1 & 83.4 & 88.6 & 93.6 & 94.7 & 97.3 & 69.3 & 85.5 \\
        \cmidrule(lr){1-12}
        \multirow{4}{*}{\textbf{\method}} & 0.50 & 90.4 & 100.0 & 60.7 & 90.4 & 93.4 & 98.6 & 97.3 & 100.0 & 80.8 & 99.8 \\
         & 0.20 & 89.9 & 99.5 & 60.2 & 89.5 & 93.4 & 98.6 & 97.3 & 100.0 & 79.4 & 98.0 \\
         & 0.10 & 90.0 & 99.6 & 60.0 & 89.3 & 92.8 & 98.0 & 97.3 & 100.0 & 77.1 & 95.2 \\
         & 0.05 & 88.4 & 97.8 & 58.6 & 87.3 & 91.3 & 96.4 & 96.6 & 99.3 & 73.1 & 90.3 \\
        \bottomrule
    
    \end{tabular*}

    \caption{Per-dataset results on ViDoRe v1 with Nemotron ColEmbed-3B v2. Each dataset reports NDCG@5 and retention (\%) relative to its Full Index. Benchmark-level macro-averages are reported in Table~\ref{tab:complete_benchmark_results}.}
    \label{tab:vidore_v1_nemotron_complete}

\end{table*}

\begin{table*}[t]
    \ContinuedFloat
    \centering
    \small
    \setlength{\tabcolsep}{2.4pt}
    \renewcommand{\arraystretch}{1.04}
    \begin{tabular*}{\textwidth}{@{\extracolsep{\fill}}cc*{5}{cc}@{}}
        \toprule
        Method & $\gamma$ & \multicolumn{2}{c}{\shortstack{Shift\\Project}} & \multicolumn{2}{c}{\shortstack{Syn.\\AI}} & \multicolumn{2}{c}{\shortstack{Syn.\\Energy}} & \multicolumn{2}{c}{\shortstack{Syn.\\Govt}} & \multicolumn{2}{c}{\shortstack{Syn.\\Health}} \\
        \cmidrule(lr){3-4}\cmidrule(lr){5-6}\cmidrule(lr){7-8}\cmidrule(lr){9-10}\cmidrule(lr){11-12}
        & & NDCG & Ret. & NDCG & Ret. & NDCG & Ret. & NDCG & Ret. & NDCG & Ret. \\
        \midrule
        Full Index & -- & 92.0 & 100.0 & 100.0 & 100.0 & 98.0 & 100.0 & 98.0 & 100.0 & 98.9 & 100.0 \\
        \midrule
        \multirow{4}{*}{1D-Pooling} & 0.50 & 92.8 & 100.8 & 99.6 & 99.6 & 96.1 & 98.0 & 98.3 & 100.3 & 99.3 & 100.4 \\
         & 0.20 & 91.1 & 99.1 & 98.3 & 98.3 & 96.8 & 98.7 & 98.0 & 100.0 & 98.9 & 100.0 \\
         & 0.10 & 88.7 & 96.4 & 94.6 & 94.6 & 96.4 & 98.3 & 97.0 & 99.0 & 98.5 & 99.6 \\
         & 0.05 & 86.5 & 94.0 & 92.6 & 92.6 & 95.0 & 96.9 & 96.6 & 98.5 & 97.0 & 98.1 \\
        \cmidrule(lr){1-12}
        \multirow{4}{*}{Semantic Clust.} & 0.50 & 92.2 & 100.2 & 100.0 & 100.0 & 97.3 & 99.2 & 98.3 & 100.3 & 98.9 & 100.0 \\
         & 0.20 & 91.2 & 99.1 & 100.0 & 100.0 & 97.1 & 99.1 & 98.4 & 100.4 & 98.5 & 99.6 \\
         & 0.10 & 89.9 & 97.7 & 99.6 & 99.6 & 97.1 & 99.0 & 97.5 & 99.5 & 97.7 & 98.7 \\
         & 0.05 & 85.8 & 93.2 & 97.3 & 97.3 & 97.1 & 99.0 & 96.8 & 98.8 & 96.3 & 97.4 \\
        \cmidrule(lr){1-12}
        \multirow{4}{*}{DocPruner} & 0.50 & 90.6 & 98.5 & 100.0 & 100.0 & 97.1 & 99.1 & 98.7 & 100.7 & 98.9 & 100.0 \\
         & 0.20 & 87.9 & 95.5 & 99.3 & 99.3 & 95.2 & 97.2 & 98.8 & 100.8 & 98.4 & 99.5 \\
         & 0.10 & 86.4 & 93.9 & 97.1 & 97.1 & 94.4 & 96.4 & 97.2 & 99.1 & 97.0 & 98.1 \\
         & 0.05 & 79.4 & 86.3 & 94.7 & 94.7 & 90.3 & 92.2 & 92.2 & 94.1 & 97.2 & 98.2 \\
        \cmidrule(lr){1-12}
        \multirow{4}{*}{SAP} & 0.50 & 92.5 & 100.5 & 100.0 & 100.0 & 97.5 & 99.5 & 98.7 & 100.7 & 99.3 & 100.4 \\
         & 0.20 & 92.4 & 100.4 & 99.6 & 99.6 & 97.6 & 99.6 & 98.2 & 100.2 & 98.2 & 99.2 \\
         & 0.10 & 91.4 & 99.4 & 99.3 & 99.3 & 96.6 & 98.6 & 98.1 & 100.2 & 98.2 & 99.2 \\
         & 0.05 & 89.1 & 96.9 & 97.5 & 97.5 & 93.9 & 95.8 & 95.3 & 97.3 & 98.2 & 99.2 \\
        \cmidrule(lr){1-12}
        \multirow{4}{*}{\textbf{\method}} & 0.50 & 92.8 & 100.8 & 100.0 & 100.0 & 97.1 & 99.1 & 98.7 & 100.7 & 99.3 & 100.4 \\
         & 0.20 & 91.8 & 99.7 & 100.0 & 100.0 & 97.1 & 99.1 & 98.7 & 100.7 & 98.9 & 100.0 \\
         & 0.10 & 90.6 & 98.5 & 100.0 & 100.0 & 96.6 & 98.6 & 99.1 & 101.2 & 99.3 & 100.4 \\
         & 0.05 & 87.6 & 95.2 & 100.0 & 100.0 & 96.3 & 98.2 & 97.5 & 99.5 & 97.0 & 98.1 \\
        \bottomrule
    
    \end{tabular*}

    \caption[]{Per-dataset results on ViDoRe v1 with Nemotron ColEmbed-3B v2 (continued).}
\end{table*}

\subsection{Detailed Results on ViDoRe v2}
\label{app:vidore_v2_complete_results}

Tables~\ref{tab:vidore_v2_colpali_complete},
\ref{tab:vidore_v2_colqwen25_complete}, and
\ref{tab:vidore_v2_nemotron_complete} report the ViDoRe v2 per-dataset results
for ColPali v1.3, ColQwen2.5, and Nemotron ColEmbed-3B v2, respectively. In
each table, the Full Index row provides the dataset-specific reference NDCG@5
used to compute retention.


\begin{table*}[p]
    \centering
    \small
    \setlength{\tabcolsep}{3.6pt}
    \renewcommand{\arraystretch}{0.96}

    \begin{tabular*}{0.90\textwidth}{@{\extracolsep{\fill}}cc*{4}{cc}@{}}
        \toprule
        Method & $\gamma$
        & \multicolumn{2}{c}{\shortstack{ESG\\Reports}}
        & \multicolumn{2}{c}{\shortstack{Biomedical\\Lectures}}
        & \multicolumn{2}{c}{\shortstack{Economics\\Reports}}
        & \multicolumn{2}{c}{\shortstack{ESG\\Human}} \\
        \cmidrule(lr){3-4}
        \cmidrule(lr){5-6}
        \cmidrule(lr){7-8}
        \cmidrule(lr){9-10}
        & & NDCG & Ret. & NDCG & Ret. & NDCG & Ret. & NDCG & Ret. \\
        \midrule

        Full Index & -- & 54.4 & 100.0 & 55.5 & 100.0 & 49.2 & 100.0 & 58.5 & 100.0 \\
        \midrule

        \multirow{4}{*}{1D-Pooling}
        & 0.50 & 45.7 & 84.0 & 53.2 & 95.8 & 49.0 & 99.7 & 48.0 & 82.1 \\
        & 0.20 & 35.3 & 65.0 & 48.6 & 87.6 & 48.1 & 97.8 & 39.5 & 67.6 \\
        & 0.10 & 27.1 & 49.8 & 44.4 & 80.0 & 44.4 & 90.2 & 26.9 & 46.0 \\
        & 0.05 & 22.3 & 41.0 & 40.1 & 72.2 & 42.6 & 86.7 & 22.8 & 39.1 \\
        \cmidrule(lr){1-10}

        \multirow{4}{*}{Semantic Clust.}
        & 0.50 & 51.7 & 95.0 & 55.1 & 99.3 & 48.1 & 97.9 & 57.9 & 99.0 \\
        & 0.20 & 45.7 & 84.0 & 52.9 & 95.3 & 45.7 & 93.0 & 52.4 & 89.6 \\
        & 0.10 & 42.5 & 78.2 & 51.2 & 92.3 & 40.8 & 82.9 & 44.5 & 76.0 \\
        & 0.05 & 37.5 & 69.1 & 48.1 & 86.7 & 38.6 & 78.5 & 38.7 & 66.2 \\
        \cmidrule(lr){1-10}

        \multirow{4}{*}{DocPruner}
        & 0.50 & 47.7 & 87.7 & 52.1 & 93.9 & 46.0 & 93.6 & 50.0 & 85.5 \\
        & 0.20 & 47.2 & 86.8 & 47.4 & 85.4 & 43.3 & 88.0 & 50.2 & 85.8 \\
        & 0.10 & 42.2 & 77.6 & 40.6 & 73.1 & 43.5 & 88.5 & 39.5 & 67.5 \\
        & 0.05 & 28.2 & 51.9 & 35.6 & 64.2 & 44.0 & 89.4 & 27.5 & 47.0 \\
        \cmidrule(lr){1-10}

        \multirow{4}{*}{SAP}
        & 0.50 & 53.5 & 98.4 & 55.2 & 99.5 & 48.8 & 99.1 & 57.4 & 98.2 \\
        & 0.20 & 46.6 & 85.8 & 53.0 & 95.6 & 44.0 & 89.5 & 50.7 & 86.7 \\
        & 0.10 & 45.6 & 83.9 & 50.4 & 90.8 & 43.8 & 89.0 & 51.4 & 87.9 \\
        & 0.05 & 41.4 & 76.1 & 43.9 & 79.1 & 37.6 & 76.4 & 47.4 & 81.1 \\
        \cmidrule(lr){1-10}

        \multirow{4}{*}{\textbf{\method}}
        & 0.50 & 55.0 & 101.2 & 55.6 & 100.1 & 48.8 & 99.2 & 56.4 & 96.4 \\
        & 0.20 & 53.1 & 97.6 & 54.6 & 98.4 & 47.9 & 97.3 & 56.0 & 95.8 \\
        & 0.10 & 50.2 & 92.4 & 53.6 & 96.6 & 45.5 & 92.5 & 51.3 & 87.7 \\
        & 0.05 & 45.9 & 84.5 & 50.8 & 91.5 & 43.3 & 88.0 & 49.3 & 84.4 \\
        \bottomrule
    \end{tabular*}

    \caption{Per-dataset results on ViDoRe v2 with ColPali v1.3.
    Each dataset reports NDCG@5 and retention (\%) relative to its Full Index.
    Benchmark-level macro-averages are reported in
    Table~\ref{tab:complete_benchmark_results}.}
    \label{tab:vidore_v2_colpali_complete}

    \vspace{8pt}

    \begin{tabular*}{0.90\textwidth}{@{\extracolsep{\fill}}cc*{4}{cc}@{}}
        \toprule
        Method & $\gamma$
        & \multicolumn{2}{c}{\shortstack{ESG\\Reports}}
        & \multicolumn{2}{c}{\shortstack{Biomedical\\Lectures}}
        & \multicolumn{2}{c}{\shortstack{Economics\\Reports}}
        & \multicolumn{2}{c}{\shortstack{ESG\\Human}} \\
        \cmidrule(lr){3-4}
        \cmidrule(lr){5-6}
        \cmidrule(lr){7-8}
        \cmidrule(lr){9-10}
        & & NDCG & Ret. & NDCG & Ret. & NDCG & Ret. & NDCG & Ret. \\
        \midrule

        Full Index & -- & 61.2 & 100.0 & 61.7 & 100.0 & 55.4 & 100.0 & 67.0 & 100.0 \\
        \midrule

        \multirow{4}{*}{1D-Pooling}
        & 0.50 & 54.9 & 89.8 & 58.5 & 94.8 & 51.0 & 92.1 & 55.3 & 82.5 \\
        & 0.20 & 38.0 & 62.1 & 53.3 & 86.3 & 47.5 & 85.7 & 35.5 & 53.0 \\
        & 0.10 & 28.0 & 45.7 & 47.5 & 77.0 & 45.6 & 82.4 & 29.1 & 43.4 \\
        & 0.05 & 21.3 & 34.8 & 43.9 & 71.1 & 42.0 & 75.8 & 24.8 & 37.0 \\
        \cmidrule(lr){1-10}

        \multirow{4}{*}{Semantic Clust.}
        & 0.50 & 56.6 & 92.5 & 60.0 & 97.2 & 52.1 & 94.1 & 64.8 & 96.8 \\
        & 0.20 & 48.0 & 78.5 & 58.7 & 95.1 & 49.0 & 88.5 & 56.6 & 84.5 \\
        & 0.10 & 42.6 & 69.7 & 55.0 & 89.0 & 44.4 & 80.2 & 49.8 & 74.4 \\
        & 0.05 & 31.3 & 51.3 & 52.1 & 84.4 & 42.0 & 75.8 & 42.2 & 63.0 \\
        \cmidrule(lr){1-10}

        \multirow{4}{*}{DocPruner}
        & 0.50 & 61.6 & 100.7 & 58.8 & 95.3 & 55.4 & 100.1 & 66.9 & 99.9 \\
        & 0.20 & 59.1 & 96.7 & 49.0 & 79.4 & 49.0 & 88.4 & 51.3 & 76.6 \\
        & 0.10 & 56.7 & 92.7 & 45.8 & 74.2 & 44.1 & 79.6 & 46.7 & 69.7 \\
        & 0.05 & 37.2 & 60.8 & 40.5 & 65.6 & 33.9 & 61.2 & 31.4 & 46.9 \\
        \cmidrule(lr){1-10}

        \multirow{4}{*}{SAP}
        & 0.50 & 60.5 & 98.9 & 61.1 & 99.0 & 53.6 & 96.8 & 64.1 & 95.7 \\
        & 0.20 & 59.6 & 97.4 & 58.9 & 95.4 & 52.3 & 94.4 & 57.8 & 86.3 \\
        & 0.10 & 53.2 & 87.0 & 55.9 & 90.6 & 52.7 & 95.1 & 53.1 & 79.3 \\
        & 0.05 & 51.3 & 83.9 & 53.5 & 86.6 & 51.9 & 93.7 & 43.8 & 65.4 \\
        \cmidrule(lr){1-10}

        \multirow{4}{*}{\textbf{\method}}
        & 0.50 & 59.9 & 98.0 & 61.4 & 99.4 & 54.9 & 99.2 & 68.0 & 101.5 \\
        & 0.20 & 59.7 & 97.7 & 59.7 & 96.7 & 54.7 & 98.8 & 67.1 & 100.2 \\
        & 0.10 & 59.3 & 97.0 & 58.0 & 93.9 & 53.7 & 97.0 & 58.9 & 87.9 \\
        & 0.05 & 56.8 & 92.9 & 55.7 & 90.2 & 48.0 & 86.8 & 58.1 & 86.7 \\
        \bottomrule
    \end{tabular*}

    \caption{Per-dataset results on ViDoRe v2 with ColQwen2.5.
    Each dataset reports NDCG@5 and retention (\%) relative to its Full Index.
    Benchmark-level macro-averages are reported in
    Table~\ref{tab:complete_benchmark_results}.}
    \label{tab:vidore_v2_colqwen25_complete}

\end{table*}


\subsection{Detailed Results on REAL-MM-RAG}
\label{app:real_mm_rag_complete_results}

Tables~\ref{tab:real_mm_rag_colpali_complete},
\ref{tab:real_mm_rag_colqwen25_complete}, and
\ref{tab:real_mm_rag_nemotron_complete} report the REAL-MM-RAG per-dataset
results for ColPali v1.3, ColQwen2.5, and Nemotron ColEmbed-3B v2,
respectively. In each table, the Full Index row provides the dataset-specific
reference NDCG@5 used to compute retention.


\begin{table*}[p]
    \centering
    \small
    \setlength{\tabcolsep}{3.6pt}
    \renewcommand{\arraystretch}{0.96}

    \begin{tabular*}{0.90\textwidth}{@{\extracolsep{\fill}}cc*{4}{cc}@{}}
        \toprule
        Method & $\gamma$
        & \multicolumn{2}{c}{\shortstack{ESG\\Reports}}
        & \multicolumn{2}{c}{\shortstack{Biomedical\\Lectures}}
        & \multicolumn{2}{c}{\shortstack{Economics\\Reports}}
        & \multicolumn{2}{c}{\shortstack{ESG\\Human}} \\
        \cmidrule(lr){3-4}
        \cmidrule(lr){5-6}
        \cmidrule(lr){7-8}
        \cmidrule(lr){9-10}
        & & NDCG & Ret. & NDCG & Ret. & NDCG & Ret. & NDCG & Ret. \\
        \midrule

        Full Index & -- & 58.5 & 100.0 & 64.9 & 100.0 & 56.7 & 100.0 & 73.4 & 100.0 \\
        \midrule

        \multirow{4}{*}{1D-Pooling}
        & 0.50 & 55.9 & 95.5 & 61.4 & 94.6 & 55.5 & 97.9 & 70.4 & 96.0 \\
        & 0.20 & 48.3 & 82.6 & 60.3 & 92.9 & 48.6 & 85.7 & 64.8 & 88.3 \\
        & 0.10 & 45.5 & 77.8 & 58.2 & 89.7 & 39.2 & 69.2 & 60.3 & 82.2 \\
        & 0.05 & 42.9 & 73.3 & 55.4 & 85.3 & 32.0 & 56.5 & 56.8 & 77.4 \\
        \cmidrule(lr){1-10}

        \multirow{4}{*}{Semantic Clust.}
        & 0.50 & 56.0 & 95.8 & 62.5 & 96.3 & 56.6 & 99.8 & 70.9 & 96.7 \\
        & 0.20 & 55.5 & 94.8 & 61.2 & 94.3 & 51.8 & 91.3 & 69.3 & 94.5 \\
        & 0.10 & 51.7 & 88.3 & 60.5 & 93.2 & 45.5 & 80.2 & 69.1 & 94.1 \\
        & 0.05 & 47.0 & 80.3 & 59.2 & 91.3 & 41.1 & 72.6 & 62.5 & 85.1 \\
        \cmidrule(lr){1-10}

        \multirow{4}{*}{DocPruner}
        & 0.50 & 58.0 & 99.1 & 62.6 & 96.4 & 55.5 & 97.9 & 72.4 & 98.6 \\
        & 0.20 & 57.6 & 98.5 & 62.0 & 95.5 & 53.8 & 94.9 & 71.2 & 97.0 \\
        & 0.10 & 56.1 & 95.9 & 59.4 & 91.5 & 50.6 & 89.3 & 69.4 & 94.5 \\
        & 0.05 & 51.8 & 88.5 & 54.0 & 83.3 & 47.0 & 82.9 & 64.4 & 87.7 \\
        \cmidrule(lr){1-10}

        \multirow{4}{*}{SAP}
        & 0.50 & 57.8 & 98.8 & 62.9 & 97.0 & 56.0 & 98.7 & 70.9 & 96.6 \\
        & 0.20 & 54.7 & 93.5 & 61.9 & 95.3 & 50.3 & 88.6 & 73.0 & 99.5 \\
        & 0.10 & 57.2 & 97.8 & 61.6 & 94.9 & 48.3 & 85.2 & 70.3 & 95.7 \\
        & 0.05 & 53.2 & 90.9 & 60.1 & 92.6 & 44.8 & 79.0 & 64.8 & 88.3 \\
        \cmidrule(lr){1-10}

        \multirow{4}{*}{\textbf{\method}}
        & 0.50 & 58.6 & 100.1 & 63.0 & 97.1 & 58.8 & 103.6 & 71.6 & 97.6 \\
        & 0.20 & 57.3 & 97.9 & 62.8 & 96.7 & 57.2 & 100.9 & 72.8 & 99.2 \\
        & 0.10 & 57.5 & 98.4 & 62.8 & 96.8 & 56.4 & 99.4 & 72.5 & 98.7 \\
        & 0.05 & 58.0 & 99.2 & 62.2 & 95.9 & 51.7 & 91.2 & 72.9 & 99.3 \\
        \bottomrule
    \end{tabular*}

    \caption{Per-dataset results on ViDoRe v2 with Nemotron ColEmbed-3B v2.
    Each dataset reports NDCG@5 and retention (\%) relative to its Full Index.
    Benchmark-level macro-averages are reported in
    Table~\ref{tab:complete_benchmark_results}.}
    \label{tab:vidore_v2_nemotron_complete}

    \vspace{8pt}

    \begin{tabular*}{0.90\textwidth}{@{\extracolsep{\fill}}cc*{4}{cc}@{}}
        \toprule
        Method & $\gamma$
        & \multicolumn{2}{c}{\shortstack{Fin\\Report}}
        & \multicolumn{2}{c}{\shortstack{Fin\\Slides}}
        & \multicolumn{2}{c}{\shortstack{Tech\\Report}}
        & \multicolumn{2}{c}{\shortstack{Tech\\Slides}} \\
        \cmidrule(lr){3-4}
        \cmidrule(lr){5-6}
        \cmidrule(lr){7-8}
        \cmidrule(lr){9-10}
        & & NDCG & Ret. & NDCG & Ret. & NDCG & Ret. & NDCG & Ret. \\
        \midrule

        Full Index & -- & 40.1 & 100.0 & 35.3 & 100.0 & 60.9 & 100.0 & 78.0 & 100.0 \\
        \midrule

        \multirow{4}{*}{1D-Pooling}
        & 0.50 & 32.5 & 80.9 & 26.7 & 75.7 & 56.1 & 92.2 & 75.7 & 97.1 \\
        & 0.20 & 23.6 & 58.9 & 16.9 & 47.9 & 50.0 & 82.1 & 68.2 & 87.4 \\
        & 0.10 & 19.0 & 47.3 & 11.7 & 33.2 & 39.5 & 64.9 & 61.2 & 78.5 \\
        & 0.05 & 15.7 & 39.1 & 7.6 & 21.6 & 31.0 & 50.9 & 56.2 & 72.0 \\
        \cmidrule(lr){1-10}

        \multirow{4}{*}{Semantic Clust.}
        & 0.50 & 39.0 & 97.2 & 33.5 & 94.9 & 60.1 & 98.7 & 77.8 & 99.7 \\
        & 0.20 & 33.2 & 82.7 & 30.5 & 86.4 & 56.2 & 92.3 & 75.9 & 97.3 \\
        & 0.10 & 28.8 & 71.8 & 27.3 & 77.3 & 50.5 & 83.0 & 72.9 & 93.4 \\
        & 0.05 & 23.9 & 59.5 & 21.1 & 59.7 & 44.4 & 72.9 & 68.6 & 88.0 \\
        \cmidrule(lr){1-10}

        \multirow{4}{*}{DocPruner}
        & 0.50 & 31.9 & 79.6 & 19.4 & 55.1 & 52.1 & 85.6 & 72.0 & 92.3 \\
        & 0.20 & 27.6 & 68.8 & 17.4 & 49.4 & 50.0 & 82.1 & 68.8 & 88.2 \\
        & 0.10 & 20.0 & 49.8 & 11.7 & 33.0 & 41.0 & 67.3 & 61.0 & 78.2 \\
        & 0.05 & 16.6 & 41.3 & 8.4 & 23.7 & 32.5 & 53.5 & 51.5 & 66.0 \\
        \cmidrule(lr){1-10}

        \multirow{4}{*}{SAP}
        & 0.50 & 38.9 & 97.0 & 33.9 & 96.0 & 59.4 & 97.6 & 77.7 & 99.5 \\
        & 0.20 & 35.8 & 89.3 & 30.4 & 86.1 & 54.9 & 90.2 & 74.8 & 95.9 \\
        & 0.10 & 32.1 & 80.0 & 24.0 & 68.1 & 50.2 & 82.5 & 71.1 & 91.2 \\
        & 0.05 & 26.2 & 65.3 & 16.1 & 45.6 & 43.0 & 70.7 & 65.6 & 84.1 \\
        \cmidrule(lr){1-10}

        \multirow{4}{*}{\textbf{\method}}
        & 0.50 & 39.9 & 99.6 & 34.8 & 98.6 & 60.6 & 99.6 & 78.3 & 100.4 \\
        & 0.20 & 39.4 & 98.3 & 32.3 & 91.4 & 58.5 & 96.1 & 77.6 & 99.5 \\
        & 0.10 & 36.6 & 91.3 & 28.8 & 81.6 & 55.3 & 90.8 & 75.3 & 96.5 \\
        & 0.05 & 33.3 & 83.0 & 21.9 & 62.1 & 50.9 & 83.7 & 70.9 & 90.9 \\
        \bottomrule
    \end{tabular*}

    \caption{Per-dataset results on REAL-MM-RAG with ColPali v1.3.
    Each dataset reports NDCG@5 and retention (\%) relative to its Full Index.
    Benchmark-level macro-averages are reported in
    Table~\ref{tab:complete_benchmark_results}.}
    \label{tab:real_mm_rag_colpali_complete}

\end{table*}


\begin{table*}[p]
    \centering
    \small
    \setlength{\tabcolsep}{3.6pt}
    \renewcommand{\arraystretch}{0.96}

    \begin{tabular*}{0.90\textwidth}{@{\extracolsep{\fill}}cc*{4}{cc}@{}}
        \toprule
        Method & $\gamma$
        & \multicolumn{2}{c}{\shortstack{Fin\\Report}}
        & \multicolumn{2}{c}{\shortstack{Fin\\Slides}}
        & \multicolumn{2}{c}{\shortstack{Tech\\Report}}
        & \multicolumn{2}{c}{\shortstack{Tech\\Slides}} \\
        \cmidrule(lr){3-4}
        \cmidrule(lr){5-6}
        \cmidrule(lr){7-8}
        \cmidrule(lr){9-10}
        & & NDCG & Ret. & NDCG & Ret. & NDCG & Ret. & NDCG & Ret. \\
        \midrule

        Full Index & -- & 53.7 & 100.0 & 45.4 & 100.0 & 73.4 & 100.0 & 83.5 & 100.0 \\
        \midrule

        \multirow{4}{*}{1D-Pooling}
        & 0.50 & 43.9 & 81.7 & 38.1 & 83.8 & 69.2 & 94.3 & 81.4 & 97.6 \\
        & 0.20 & 31.0 & 57.7 & 24.2 & 53.3 & 57.6 & 78.5 & 73.7 & 88.3 \\
        & 0.10 & 27.5 & 51.2 & 19.2 & 42.2 & 48.6 & 66.3 & 65.6 & 78.6 \\
        & 0.05 & 20.4 & 38.0 & 13.5 & 29.8 & 41.9 & 57.1 & 59.5 & 71.3 \\
        \cmidrule(lr){1-10}

        \multirow{4}{*}{Semantic Clust.}
        & 0.50 & 50.7 & 94.4 & 45.0 & 98.9 & 71.8 & 97.8 & 83.1 & 99.5 \\
        & 0.20 & 45.5 & 84.7 & 39.2 & 86.3 & 67.4 & 91.8 & 81.9 & 98.1 \\
        & 0.10 & 38.7 & 72.0 & 34.2 & 75.2 & 60.8 & 82.8 & 78.7 & 94.3 \\
        & 0.05 & 30.4 & 56.6 & 24.3 & 53.4 & 50.4 & 68.7 & 73.5 & 88.1 \\
        \cmidrule(lr){1-10}

        \multirow{4}{*}{DocPruner}
        & 0.50 & 50.4 & 93.8 & 31.0 & 68.2 & 69.1 & 94.1 & 79.0 & 94.7 \\
        & 0.20 & 37.6 & 70.0 & 13.3 & 29.2 & 49.6 & 67.6 & 63.0 & 75.5 \\
        & 0.10 & 32.5 & 60.5 & 9.7 & 21.3 & 42.1 & 57.4 & 59.3 & 71.1 \\
        & 0.05 & 20.7 & 38.5 & 8.5 & 18.7 & 34.1 & 46.5 & 54.3 & 65.1 \\
        \cmidrule(lr){1-10}

        \multirow{4}{*}{SAP}
        & 0.50 & 53.6 & 99.9 & 42.6 & 93.8 & 71.5 & 97.5 & 83.2 & 99.6 \\
        & 0.20 & 50.0 & 93.2 & 35.4 & 78.0 & 63.2 & 86.0 & 80.0 & 95.8 \\
        & 0.10 & 42.9 & 79.9 & 28.1 & 61.7 & 54.7 & 74.6 & 75.5 & 90.5 \\
        & 0.05 & 32.1 & 59.7 & 20.0 & 43.9 & 45.2 & 61.6 & 69.6 & 83.5 \\
        \cmidrule(lr){1-10}

        \multirow{4}{*}{\textbf{\method}}
        & 0.50 & 53.9 & 100.4 & 44.3 & 97.6 & 72.2 & 98.4 & 83.2 & 99.7 \\
        & 0.20 & 52.8 & 98.4 & 41.8 & 91.9 & 68.8 & 93.7 & 81.8 & 98.1 \\
        & 0.10 & 50.2 & 93.5 & 36.8 & 81.0 & 63.2 & 86.1 & 78.9 & 94.6 \\
        & 0.05 & 44.6 & 83.0 & 29.0 & 63.8 & 56.6 & 77.1 & 74.1 & 88.8 \\
        \bottomrule
    \end{tabular*}

    \caption{Per-dataset results on REAL-MM-RAG with ColQwen2.5.
    Each dataset reports NDCG@5 and retention (\%) relative to its Full Index.
    Benchmark-level macro-averages are reported in
    Table~\ref{tab:complete_benchmark_results}.}
    \label{tab:real_mm_rag_colqwen25_complete}

    \vspace{8pt}

    \begin{tabular*}{0.90\textwidth}{@{\extracolsep{\fill}}cc*{4}{cc}@{}}
        \toprule
        Method & $\gamma$
        & \multicolumn{2}{c}{\shortstack{Fin\\Report}}
        & \multicolumn{2}{c}{\shortstack{Fin\\Slides}}
        & \multicolumn{2}{c}{\shortstack{Tech\\Report}}
        & \multicolumn{2}{c}{\shortstack{Tech\\Slides}} \\
        \cmidrule(lr){3-4}
        \cmidrule(lr){5-6}
        \cmidrule(lr){7-8}
        \cmidrule(lr){9-10}
        & & NDCG & Ret. & NDCG & Ret. & NDCG & Ret. & NDCG & Ret. \\
        \midrule

        Full Index & -- & 69.5 & 100.0 & 61.5 & 100.0 & 77.4 & 100.0 & 86.6 & 100.0 \\
        \midrule

        \multirow{4}{*}{1D-Pooling}
        & 0.50 & 65.7 & 94.4 & 60.2 & 97.8 & 75.7 & 97.9 & 86.0 & 99.3 \\
        & 0.20 & 56.2 & 80.8 & 54.6 & 88.8 & 71.8 & 92.7 & 83.5 & 96.4 \\
        & 0.10 & 43.1 & 62.0 & 48.7 & 79.2 & 65.2 & 84.3 & 78.1 & 90.1 \\
        & 0.05 & 34.2 & 49.2 & 44.3 & 72.0 & 57.0 & 73.6 & 74.4 & 85.9 \\
        \cmidrule(lr){1-10}

        \multirow{4}{*}{Semantic Clust.}
        & 0.50 & 67.5 & 97.0 & 60.9 & 99.0 & 76.4 & 98.8 & 86.4 & 99.8 \\
        & 0.20 & 62.7 & 90.1 & 60.2 & 97.8 & 74.5 & 96.2 & 85.8 & 99.1 \\
        & 0.10 & 54.9 & 78.9 & 56.1 & 91.3 & 71.6 & 92.5 & 84.4 & 97.5 \\
        & 0.05 & 46.0 & 66.2 & 51.2 & 83.2 & 64.9 & 83.9 & 81.7 & 94.4 \\
        \cmidrule(lr){1-10}

        \multirow{4}{*}{DocPruner}
        & 0.50 & 68.2 & 98.1 & 60.6 & 98.5 & 76.3 & 98.6 & 86.8 & 100.2 \\
        & 0.20 & 62.4 & 89.7 & 56.4 & 91.6 & 72.1 & 93.2 & 84.8 & 97.9 \\
        & 0.10 & 56.4 & 81.1 & 48.1 & 78.3 & 64.3 & 83.1 & 82.4 & 95.1 \\
        & 0.05 & 44.5 & 63.9 & 39.4 & 64.0 & 53.6 & 69.3 & 78.3 & 90.4 \\
        \cmidrule(lr){1-10}

        \multirow{4}{*}{SAP}
        & 0.50 & 69.0 & 99.2 & 60.9 & 99.0 & 76.9 & 99.3 & 86.6 & 100.0 \\
        & 0.20 & 66.5 & 95.6 & 59.6 & 97.0 & 74.7 & 96.5 & 85.8 & 99.1 \\
        & 0.10 & 63.0 & 90.5 & 58.5 & 95.1 & 72.0 & 93.1 & 84.9 & 98.0 \\
        & 0.05 & 58.1 & 83.6 & 53.9 & 87.6 & 66.6 & 86.1 & 82.1 & 94.8 \\
        \cmidrule(lr){1-10}

        \multirow{4}{*}{\textbf{\method}}
        & 0.50 & 69.3 & 99.7 & 61.7 & 100.4 & 77.5 & 100.1 & 86.6 & 99.9 \\
        & 0.20 & 68.6 & 98.6 & 61.3 & 99.7 & 76.8 & 99.2 & 86.4 & 99.8 \\
        & 0.10 & 66.6 & 95.7 & 60.9 & 99.0 & 74.7 & 96.6 & 86.6 & 100.0 \\
        & 0.05 & 64.8 & 93.2 & 59.2 & 96.3 & 72.7 & 94.0 & 85.4 & 98.6 \\
        \bottomrule
    \end{tabular*}

    \caption{Per-dataset results on REAL-MM-RAG with Nemotron ColEmbed-3B v2.
    Each dataset reports NDCG@5 and retention (\%) relative to its Full Index.
    Benchmark-level macro-averages are reported in
    Table~\ref{tab:complete_benchmark_results}.}
    \label{tab:real_mm_rag_nemotron_complete}

\end{table*}

\FloatBarrier


\section{Detailed Comparison with Trained Compression}
\label{app:trained_compression_comparison}

\subsection{Controlled Comparison Protocol}
\label{app:trained_compression_protocol}

We compare \method with AGC, a trained compression method that learns
universal query tokens for saliency-guided representative selection and
weighted aggregation \citep{qin2026agc}. AGC uses last-layer attention
from the learned tokens to select representative document vectors, assigns
the remaining vectors to their most similar representatives, and summarizes
each resulting group through attention-weighted aggregation.

We reproduce AGC within the same codebase used for \method and use the same
bidirectional ColQwen2.5 backbone setup. AGC is trained using the same data
and LoRA recipe as the ColQwen2.5 retriever described in
Appendix~\ref{app:retrieval_backbone_details}, while jointly learning
128 universal query tokens. In contrast, \method is applied post hoc to
the corresponding retrieval checkpoint, introduces no additional parameters,
and requires no additional training. We use the calibrated ColQwen2.5
configuration from Appendix~\ref{app:anchorfold_configuration}.

Unlike the retention-ratio experiments in the main evaluation, this
comparison uses a common fixed document-vector cap. For a page with \(N\)
valid document vectors, both methods use
\begin{equation}
    m=\min(N,128).
\end{equation}
All other preprocessing and retrieval settings are held fixed, including
the image processor, document corpora, test queries, relevance judgments,
query encoding, and MaxSim scoring. Evaluation follows
Appendix~\ref{app:benchmark_evaluation_details} over the same 18 constituent
datasets.

\subsection{Complete Comparison Results}
\label{app:trained_compression_complete_results}

Table~\ref{tab:trained_compression_complete_results} reports the complete
comparison. At the benchmark level, \method achieves higher macro-average
NDCG@5 than AGC on all three evaluation suites, with improvements of
\(0.5\), \(2.8\), and \(1.4\) points on ViDoRe v1, ViDoRe v2, and
REAL-MM-RAG, respectively. The largest benchmark-level improvement occurs
on ViDoRe v2, where \method reaches 59.8 NDCG@5 compared with 57.0 for
AGC.

The per-dataset results show that the benchmark-level improvements are
not driven by a single collection. \method achieves higher NDCG@5 on
14 of the 18 constituent datasets and on all four REAL-MM-RAG
collections. On ViDoRe v2, the largest improvement occurs on the fully
human-labeled ESG collection, where \method exceeds AGC by 5.6 NDCG@5
points, with additional gains on ESG Reports and Economics Reports.
On REAL-MM-RAG, \method improves over AGC on both report and slide
collections, with the largest gain of 3.5 points on FinReport.

AGC remains stronger on four individual datasets: ArXivQA, InfoVQA,
Synthetic AI, and Biomedical Lectures. Nevertheless, \method obtains a
higher macro-average on each of the three evaluated benchmark suites.
These results show that \method can achieve competitive fixed-budget
retrieval performance without additional parameters or training, while
providing higher aggregate NDCG@5 under this comparison protocol.

\begin{table*}[t]
    \centering
    \small
    \setlength{\tabcolsep}{4.5pt}
    \renewcommand{\arraystretch}{1.08}
    \begin{tabular}[t]{@{}lcc@{}}
            \toprule
            Dataset
            & \shortstack{AGC\\(Trained)}
            & \shortstack{\method\\(Training-Free)} \\
            \midrule

            \multicolumn{3}{@{}l}{\textbf{ViDoRe v1} \textit{(10 datasets)}} \\
            \addlinespace[1pt]
            ArXivQA               & \textbf{89.7} & 88.9 \\
            DocVQA                & 53.6 & \textbf{53.9} \\
            InfoVQA               & \textbf{89.1} & 88.9 \\
            TabFQuAD              & 91.2 & \textbf{92.9} \\
            TAT-DQA               & 77.9 & \textbf{79.9} \\
            Shift Project         & 87.1 & \textbf{88.3} \\
            Synthetic AI          & \textbf{99.6} & 97.6 \\
            Synthetic Energy      & 93.2 & \textbf{94.7} \\
            Synthetic Government  & 97.5 & \textbf{98.5} \\
            Synthetic Healthcare  & 98.0 & \textbf{98.8} \\
            \addlinespace[1pt]
            \textit{Average}      & 87.7 & \textbf{88.2} \\
            \bottomrule
        \end{tabular}%
    \hspace{7em}
    \begin{tabular}[t]{@{}lcc@{}}
            \toprule
            Dataset
            & \shortstack{AGC\\(Trained)}
            & \shortstack{\method\\(Training-Free)} \\
            \midrule

            \multicolumn{3}{@{}l}{\textbf{ViDoRe v2} \textit{(4 datasets)}} \\
            \addlinespace[1pt]
            ESG Reports           & 56.7 & \textbf{59.4} \\
            Biomedical Lectures   & \textbf{60.1} & 59.5 \\
            Economics Reports     & 51.9 & \textbf{55.5} \\
            ESG Reports (Human)   & 59.2 & \textbf{64.8} \\
            \addlinespace[1pt]
            \textit{Average}      & 57.0 & \textbf{59.8} \\

            \midrule
            \multicolumn{3}{@{}l}{\textbf{REAL-MM-RAG} \textit{(4 datasets)}} \\
            \addlinespace[1pt]
            FinReport             & 49.3 & \textbf{52.8} \\
            FinSlides             & 39.3 & \textbf{40.0} \\
            TechReport            & 67.3 & \textbf{68.5} \\
            TechSlides            & 80.9 & \textbf{81.3} \\
            \addlinespace[1pt]
            \textit{Average}      & 59.2 & \textbf{60.6} \\
            \bottomrule
        \end{tabular}

    \caption{
    Controlled fixed-budget comparison with trained AGC using at most
    128 document vectors per page. All entries are NDCG@5 on a 0--100
    scale. Average is the unweighted mean over the constituent datasets
    of each benchmark. The better result in each row is shown in bold.
    }
    \label{tab:trained_compression_complete_results}
\end{table*}

\section{Calibration Details and Validation}
\label{app:calibration_details}

\subsection{Joint SR-Based Calibration Protocol}
\label{app:joint_sr_calibration_protocol}

We calibrate the propagation depth \(K\) and layer window
\(\mathcal{L}\) jointly using Score Retention (SR), adapted from the
label-free diagnostic introduced by SAP \citep{liu2026sap}. Let
\(\mathbf{Q}_n=[\mathbf{q}_{n,1},\ldots,\mathbf{q}_{n,M_n}]\) and
\(\mathbf{E}_n=[\mathbf{e}_{n,1},\ldots,\mathbf{e}_{n,N_n}]\) denote
the query and full document representations for calibration pair \(n\).
Their MaxSim score is
\begin{equation}
    S(\mathbf{Q}_n,\mathbf{E}_n)
    =
    \sum_{i=1}^{M_n}
    \max_{1\leq j\leq N_n}
    \mathbf{q}_{n,i}^{\top}\mathbf{e}_{n,j}.
\end{equation}
For propagation depth \(K\) and layer set \(\mathcal{L}\), let
\(\widehat{\mathbf{E}}_{n,K,\mathcal{L}}\) be the representation
produced by the complete \method operator, including anchor selection,
token assignment, and centrality-weighted aggregation. Pair-level SR is
defined as
\begin{equation}
    R_n(K,\mathcal{L})
    =
    100
    \times
    \frac{
        S(\mathbf{Q}_n,\widehat{\mathbf{E}}_{n,K,\mathcal{L}})
    }{
        S(\mathbf{Q}_n,\mathbf{E}_n)
    }.
    \label{eq:appendix_pair_sr}
\end{equation}
Unlike NDCG, SR measures the fidelity of a compressed representation
for an individual query--page pair without introducing corpus-level
ranking effects or requiring relevance judgments.

We use 500 held-out query--page pairs sampled from the ColPali training
corpus with random seed 42. The calibration pairs are disjoint from all
ViDoRe v1, ViDoRe v2, and REAL-MM-RAG evaluation splits. Calibration is
performed at \(\gamma_{\mathrm{cal}}=0.20\); a page with \(N_n\) valid
visual tokens is therefore represented using
\(m_n=\lceil0.20N_n\rceil\) vectors. We use zero-based layer indices,
a relative window width of \(\rho=0.2\), and candidate propagation
depths
\begin{equation}
    \mathcal{K}
    =
    \{2,3,4,6,8,10,12\}.
    \label{eq:appendix_candidate_depths}
\end{equation}

The calibration consists of two stages. First, for every candidate
\(K\), we apply \method using each backbone layer independently and
average pair-level SR over the calibration set, producing the
single-layer profile
\begin{equation}
    \overline{R}_{K}(l)
    =
    \frac{1}{500}
    \sum_{n=1}^{500}
    R_n(K,\{l\}),
    \qquad
    l=0,\ldots,L_{\mathrm{tot}}-1,
\end{equation}
where \(L_{\mathrm{tot}}\) denotes the total number of backbone layers.
Let \(m_K\) be the median of this profile across all layers. We define
\(l_K^{\star}\) as the beginning of the longest trailing suffix whose
SR remains below \(m_K\):
\begin{equation}
    l_K^{\star}
    =
    \min
    \left\{
        l:
        \overline{R}_{K}(j)<m_K
        \ \text{for every}\
        j\in\{l,\ldots,L_{\mathrm{tot}}-1\}
    \right\}.
\end{equation}
With
\begin{equation}
    w=\left\lceil\rho L_{\mathrm{tot}}\right\rceil,
\end{equation}
the candidate window for \(K\) is placed immediately before this
trailing low-SR region:
\begin{equation}
    \mathcal{L}_K
    =
    \left\{
        \max(0,l_K^{\star}-w),
        \ldots,
        l_K^{\star}-1
    \right\}.
    \label{eq:appendix_sr_window}
\end{equation}
This procedure yields widths of 4, 8, and 6 layers for ColPali v1.3,
ColQwen2.5, and Nemotron ColEmbed-3B v2, respectively.

Second, we evaluate each joint candidate
\((K,\mathcal{L}_K)\) using the deployment-faithful multi-layer
operator: layer-normalized centrality is averaged over
\(\mathcal{L}_K\), after which the complete focus-then-fold procedure
constructs the compressed representation. We denote the resulting
pair-level values by
\begin{equation}
    R_{n,K}
    =
    R_n(K,\mathcal{L}_K),
    \qquad
    \overline{R}_K
    =
    \frac{1}{500}
    \sum_{n=1}^{500}R_{n,K}.
\end{equation}

We select \(K\) using a paired one-standard-error rule. Let
\begin{equation}
    K_{\mathrm{best}}
    =
    \arg\max_{K\in\mathcal{K}}\overline{R}_K
\end{equation}
and define the pairwise gap
\begin{equation}
    d_n(K)
    =
    R_{n,K_{\mathrm{best}}}-R_{n,K},
\end{equation}
with mean paired gap
\begin{equation}
    \overline{d}(K)
    =
    \frac{1}{500}
    \sum_{n=1}^{500}d_n(K).
\end{equation}
A candidate is eligible when its mean paired gap does not exceed one
standard error of the paired gaps:
\begin{equation}
    \overline{d}(K)
    \leq
    \operatorname{SE}[d(K)]
    =
    \frac{
        \operatorname{sd}\!\left(
            d_1(K),\ldots,d_{500}(K)
        \right)
    }{
        \sqrt{500}
    }.
    \label{eq:appendix_paired_one_se}
\end{equation}
Among the eligible candidates, we select the smallest \(K\). The rule
uses paired differences from the same calibration examples and
therefore accounts for the strong correlation between candidate
configurations. It also avoids selecting a larger propagation depth
for a negligible improvement in mean SR.

\subsection{Joint Calibration Results}
\label{app:joint_sr_calibration_results}

Table~\ref{tab:joint_sr_calibration} reports the complete joint-SR
sweep. SR is reported on a percentage scale as the mean and standard
error over the 500 calibration pairs. Each entry is evaluated with the
layer window independently obtained from the single-layer SR profile
for that \(K\); the layer indices are zero-based and inclusive.

\begin{table*}[t]
    \centering
    \small
    \setlength{\tabcolsep}{4.0pt}
    \renewcommand{\arraystretch}{1.07}
    
    \begin{tabular*}{\textwidth}
        {@{\extracolsep{\fill}}c cc cc cc@{}}
        \toprule
        & \multicolumn{2}{c}{ColPali v1.3}
        & \multicolumn{2}{c}{ColQwen2.5}
        & \multicolumn{2}{c}{Nemotron ColEmbed-3B v2} \\
        \cmidrule(lr){2-3}
        \cmidrule(lr){4-5}
        \cmidrule(lr){6-7}
        \(K\)
        & Layers & SR (\%)
        & Layers & SR (\%)
        & Layers & SR (\%) \\
        \midrule
        2
        & 11--14 & \(95.948 \pm 0.127\)
        & 25--32 & \(95.999 \pm 0.174\)
        & 14--19 & \(99.444 \pm 0.018\) \\
        3
        & 11--14 & \(95.967 \pm 0.136\)
        & 25--32 & \(96.014 \pm 0.174\)
        & 14--19 & \(99.448 \pm 0.019\) \\
        4
        & 11--14 & \(95.981 \pm 0.137\)
        & 25--32 & \(96.005 \pm 0.175\)
        & 14--19 & \(99.450 \pm 0.019\) \\
        \textbf{6}
        & \textbf{11--14} & \(\mathbf{96.034 \pm 0.145}\)
        & \textbf{25--32} & \(\mathbf{96.043 \pm 0.188}\)
        & \textbf{14--19} & \(\mathbf{99.463 \pm 0.018}\) \\
        8
        & 11--14 & \(96.001 \pm 0.138\)
        & 25--32 & \(96.016 \pm 0.195\)
        & 14--19 & \(99.464 \pm 0.019\) \\
        10
        & 11--14 & \(95.975 \pm 0.140\)
        & 24--31 & \(96.162 \pm 0.182\)
        & 14--19 & \(99.462 \pm 0.019\) \\
        12
        & 11--14 & \(95.990 \pm 0.140\)
        & 24--31 & \(96.155 \pm 0.173\)
        & 14--19 & \(99.462 \pm 0.019\) \\
        \bottomrule
    \end{tabular*}

    \caption{
    Joint SR-based calibration across propagation depths and retrieval
    backbones. Values are mean SR (\%) \(\pm\) standard error over 500
    held-out query--page pairs. Each \(K\) is evaluated with its own
    SR-selected layer window. Bold indicates the configuration retained
    by the paired one-standard-error rule. The reported marginal standard
    errors are descriptive; eligibility is determined from paired
    per-example gaps in Equation~\ref{eq:appendix_paired_one_se}.
    }
    \label{tab:joint_sr_calibration}
\end{table*}

The highest mean SR occurs at \(K=6\) for ColPali v1.3, \(K=10\) for
ColQwen2.5, and \(K=8\) for Nemotron ColEmbed-3B v2. These maxima differ
by at most small fractions of one percentage point from the neighboring
candidates. Applying the paired rule in
Equation~\ref{eq:appendix_paired_one_se}, \(K=6\) is the smallest
eligible depth for all three backbones, whereas every candidate with
\(K<6\) falls outside the corresponding one-standard-error set. At
\(K=6\), the calibrated windows are layers 11--14, 25--32, and 14--19
for ColPali v1.3, ColQwen2.5, and Nemotron ColEmbed-3B v2,
respectively. We therefore use this common propagation depth and the
three backbone-specific windows in all reported experiments.

\subsection{SR-Guided Layer-Window Validation}
\label{app:layer_window_selection_validation}

We next validate whether the label-free SR criterion identifies a
layer region that also preserves downstream retrieval effectiveness.
Using ColQwen2.5 on the complete ViDoRe v2 benchmark, we fix
\(K=6\) and \(\gamma=0.20\), and evaluate the nine canonical
20\%-wide windows spanning the backbone. SR is computed on the held-out
calibration pairs, whereas Avg.\ NDCG@5 is the unweighted macro-average
over the four ViDoRe v2 datasets. NDCG@5 is used only for post-hoc
validation and does not participate in window selection.

\begin{table}[t]
    \centering
    \small
    \setlength{\tabcolsep}{5.5pt}
    \renewcommand{\arraystretch}{1.07}
    
    \begin{tabular}{@{}cccc@{}}
        \toprule
        Relative window (\%)
        & Layers
        & SR (\%)
        & Avg.\ NDCG@5 \\
        \midrule
        0--20   & 0--7   & 94.365 & 55.64 \\
        10--30  & 3--10  & 94.969 & 56.42 \\
        20--40  & 7--14  & 95.187 & 57.84 \\
        30--50  & 10--17 & 95.495 & 59.47 \\
        40--60  & 14--21 & 95.392 & 58.93 \\
        50--70  & 18--25 & 95.955 & 59.93 \\
        60--80  & 21--28 & 95.874 & 60.00 \\
        \textbf{70--90}
                 & \textbf{25--32}
                 & \textbf{96.043}
                 & \textbf{60.31} \\
        80--100 & 28--35 & 95.272 & 59.35 \\
        \bottomrule
    \end{tabular}

    \caption{
    Post-hoc validation of the SR-guided layer window on ViDoRe v2
    with ColQwen2.5 at \(K=6\) and \(\gamma=0.20\). Layer indices are
    zero-based and inclusive. Bold indicates the SR-selected window.
    }
    \label{tab:layer_window_validation}
\end{table}

The SR-selected 70--90\% window attains both the highest calibration
SR and the highest downstream NDCG@5. Retrieval effectiveness
generally improves from the early layers toward the middle-to-late
backbone region, but decreases when the window is moved to the final
20\% of the backbone. In particular, shifting from layers 25--32 to
layers 28--35 decreases SR from 96.043\% to 95.272\% and Avg.\
NDCG@5 from 60.31 to 59.35. This agreement suggests that the trailing low-SR region provides a useful label-free signal for avoiding layer regions whose attention patterns yield less effective compression under the fixed vector budget.

\subsection{Propagation-Depth Validation}
\label{app:propagation_depth_selection_validation}

Finally, we validate the selected propagation depth on ViDoRe v2 with
ColQwen2.5 at \(\gamma=0.20\). To preserve the joint nature of the
calibration, each \(K\) is evaluated with the layer window selected
from its own single-layer SR profile. Thus,
\(K\in\{2,3,4,6,8\}\) uses layers 25--32, whereas
\(K\in\{10,12\}\) uses layers 24--31. The benchmark evaluation is
again post-hoc and does not modify the SR-selected configuration.

\begin{table}[t]
    \centering
    \small
    \setlength{\tabcolsep}{8pt}
    \renewcommand{\arraystretch}{1.07}
    
    \begin{tabular}{@{}ccc@{}}
        \toprule
        \(K\)
        & Layers
        & Avg.\ NDCG@5 \\
        \midrule
        2  & 25--32 & 58.64 \\
        3  & 25--32 & 59.85 \\
        4  & 25--32 & 59.72 \\
        \textbf{6}
           & \textbf{25--32}
           & \textbf{60.31} \\
        8  & 25--32 & 60.10 \\
        10 & 24--31 & 60.39 \\
        12 & 24--31 & 60.42 \\
        \bottomrule
    \end{tabular}

    \caption{
    Post-hoc propagation-depth validation on ViDoRe v2 with
    ColQwen2.5 at \(\gamma=0.20\). Each depth uses its independently
    SR-selected layer window. Bold indicates the deployed
    configuration.
    }
    \label{tab:propagation_depth_validation}
\end{table}

Increasing the propagation depth from \(K=2\) to \(K=6\) improves
Avg.\ NDCG@5 by approximately 1.7 points, showing a clear benefit over
the shallowest calibrated setting. Beyond \(K=6\), performance largely
saturates: \(K=8\) is slightly lower, while \(K=10\) and \(K=12\)
provide only marginal additional improvements. These results are
consistent with the paired SR calibration: \(K=6\) captures the
substantial benefit of recursive propagation while avoiding additional
iterations whose downstream gains are small and backbone-dependent. Together with the backbone-wise paired SR results in
Table~\ref{tab:joint_sr_calibration}, this ColQwen2.5 sensitivity
analysis supports retaining \(K=6\) as the common propagation depth
for all three backbones.

\begin{table*}[t]
    \centering
    \small
    \setlength{\tabcolsep}{3.2pt}
    \renewcommand{\arraystretch}{1.08}
    
    \begin{tabular*}{\textwidth}{
        @{\extracolsep{\fill}}lcccccc@{}
    }
        \toprule
        & \multicolumn{2}{c}{Retrieval}
        & \multicolumn{4}{c}{Latency (ms/page)} \\
        \cmidrule(lr){2-3}
        \cmidrule(lr){4-7}

        Method
        & NDCG@5
        & Ret. (\%)
        & Comp.
        & \shortstack{Total\\Mean \(\pm\) SD}
        & P95
        & Index / Full \\
        \midrule

        Full Index
        & 61.3
        & 100.0
        & --
        & \(88.119 \pm 0.053\)
        & 92.047
        & 1.000 \\

        SAP
        & 53.7
        & 87.6
        & \(2.324 \pm 0.005\)
        & \(92.915 \pm 0.039\)
        & 97.116
        & 1.054 \\

        Semantic Clust.
        & 48.0
        & 78.3
        & \(29.497 \pm 0.054\)
        & \(117.367 \pm 0.091\)
        & 122.424
        & 1.332 \\

        1D-Pooling
        & 37.6
        & 61.3
        & \(0.3300 \pm 0.0001\)
        & \(88.327 \pm 0.027\)
        & 92.254
        & 1.002 \\

        DocPruner
        & 48.3
        & 78.8
        & \(0.823 \pm 0.003\)
        & \(89.084 \pm 0.064\)
        & 93.025
        & 1.011 \\

        \addlinespace[1pt]

        \textbf{\method}
        & \textbf{57.5}
        & \textbf{93.8}
        & \(7.103 \pm 0.015\)
        & \(97.726 \pm 0.017\)
        & 101.858
        & 1.109 \\

        \bottomrule
    \end{tabular*}

    \caption{
    Index-time efficiency on ViDoRe v2 using ColQwen2.5 at
    \(\gamma=0.10\). Retrieval metrics are evaluated on the complete
    ViDoRe v2 benchmark, while latency is measured on the fixed
    200-page sample. ``Comp.'' denotes compression latency, and
    Index / Full denotes mean total indexing latency normalized to
    the Full Index. Bold indicates the best retrieval result among
    the compressed methods.
    }
    \label{tab:complete_efficiency_results}

\end{table*}

\section{Efficiency Evaluation Details}
\label{app:efficiency_details}

\subsection{Efficiency Measurement Protocol}
\label{app:efficiency_measurement_protocol}

We evaluate index-time efficiency on ViDoRe v2 using the same
bidirectional ColQwen2.5 retrieval checkpoint and native image processor
as in the main experiments. All compression methods are evaluated at the
target document-vector retention ratio \(\gamma=0.10\). For \method, we
use the calibrated propagation depth \(K=6\) and the zero-indexed layer
window \(\mathcal{L}=\{25,\ldots,32\}\). The number of valid visual
tokens is capped at 768.

Latency is measured on a fixed stratified sample of 200 document pages,
with 50 pages sampled uniformly without replacement from each of the
four ViDoRe v2 subsets: ESG Reports, Biomedical Lectures, Economics
Reports, and ESG Reports (Human). We fix the sampling seed to 42
and reuse the same sample manifest for every method. Retrieval quality
reported in Table~\ref{tab:complete_efficiency_results}, in contrast, is
evaluated on the complete ViDoRe v2 benchmark rather than on the
200-page latency sample.

Each method is evaluated in a separate process on a single NVIDIA A800
80GB GPU with batch size 1 and bfloat16 inference. All sampled inputs
are prepared before timing. We exclude image I/O, image preprocessing,
and host-to-device transfer so that the measurements isolate model-side
document encoding and index construction. Each method is first run on
30 untimed warm-up pages. We then perform three complete measured passes
over the same 200-page manifest, yielding 600 per-page observations.
Wall-clock latency is measured using \texttt{time.perf\_counter}, with
CUDA synchronization immediately before and after each timed region.

We report two latency quantities. \emph{Compression latency} measures
only the method-specific compression procedure, starting after the
backbone outputs required by the compressor are available and ending
when the compressed document vectors have been constructed.
\emph{Total indexing latency} measures the complete document-side
indexing procedure from prepared device-resident model inputs to the
final document vectors.

For each latency quantity, the reported mean is computed over all 600
per-page observations. The value following \(\pm\) is the standard
deviation of the three full-pass means, while P95 is computed over all
600 per-page observations. We additionally report total indexing
latency normalized to the Full Index:
\begin{equation}
    R_{\mathrm{index}}
    =
    \frac{T_{\mathrm{method}}}{T_{\mathrm{full}}}.
\end{equation}
The corresponding indexing overhead is
\begin{equation}
    \Delta_{\mathrm{index}}
    =
    \left(
        R_{\mathrm{index}}-1
    \right)
    \times 100\%.
\end{equation}
The normalized quantity \(R_{\mathrm{index}}\) corresponds to the
horizontal axis of Figure~\ref{fig:efficiency_tradeoff} in the main
paper.

For \method and the fixed-budget baselines, a page containing \(N\)
valid document vectors is represented using
\(m=\lceil0.10N\rceil\) output vectors, following the common budget
definition in Appendix~\ref{app:baseline_details}. DocPruner instead
uses its globally calibrated pruning threshold and realizes a vector
retention of 10.75\% on the fixed latency sample. NDCG@5 retention is
computed relative to the Full Index on the complete ViDoRe v2
benchmark.

\subsection{Computational Complexity}
\label{app:efficiency_complexity}

We analyze the additional index-time computation introduced by
\method. Let \(N\) denote the number of valid visual tokens,
\(m=\lceil\gamma N\rceil\) the number of anchors, \(H\) the number of
attention heads, and \(L_s=|\mathcal{L}|\) the number of selected
layers. Let \(d\) and \(d_r\) denote the backbone hidden dimension and
retrieval embedding dimension, respectively.

In the focus stage, constructing and normalizing the visual attention
graphs over the selected layers and heads requires
\(\mathcal{O}(L_sHN^2)\) operations. Recursive Attention Propagation
performs \(K\) matrix--vector multiplications for each selected layer
and attention head, giving a focus-stage complexity of
\begin{equation}
    \mathcal{C}_{\mathrm{focus}}
    =
    \mathcal{O}\!\left(
        L_sHKN^2
    \right).
\end{equation}
The subsequent head integration and layer-wise score aggregation require
only linear computation in \(N\) and are lower-order terms.

In the fold stage, selecting the \(m\) highest-centrality anchors costs
\(\mathcal{O}(N\log m)\). Assigning visual tokens to their most similar
anchors in the \(d_r\)-dimensional normalized retrieval space requires
\(\mathcal{O}(Nmd_r)\) operations. Centrality-weighted aggregation over
the resulting anchor-centered groups costs \(\mathcal{O}(Nd)\), and
projecting the \(m\) aggregated hidden states into the retrieval space
costs \(\mathcal{O}(mdd_r)\). The overall additional index-time
complexity of \method is therefore
\begin{equation}
\begin{split}
    \mathcal{C}_{\method}
    =
    \mathcal{O}\big(
        &L_sHKN^2
        + Nmd_r \\
        &+ mdd_r
        + Nd
        + N\log m
    \big).
\end{split}
\end{equation}

Since \(m=\lceil\gamma N\rceil\), the leading terms can be expressed as
\begin{equation}
    \mathcal{O}\!\left(
        L_sHKN^2
        + \gamma N^2d_r
        + \gamma Ndd_r
    \right).
\end{equation}
For a fixed backbone and compression configuration,
\(L_s\), \(H\), \(K\), \(\gamma\), \(d\), and \(d_r\) are constants;
thus, the index-time complexity of \method is quadratic in the number
of valid visual tokens. In the evaluated ColQwen2.5 configuration,
\(K=6\), \(L_s=8\), and \(N\leq768\).

\method introduces no additional computation during query encoding.
For a query containing \(Q\) query vectors, exact per-page MaxSim
scoring decreases from
\(\mathcal{O}(QNd_r)\) for the Full Index to
\(\mathcal{O}(Qmd_r)=\mathcal{O}(\gamma QNd_r)\) after compression.
Similarly, document-index storage decreases from
\(\mathcal{O}(Nd_r)\) to
\(\mathcal{O}(md_r)=\mathcal{O}(\gamma Nd_r)\).

\subsection{Complete Efficiency Results}
\label{app:complete_efficiency_results}

Table~\ref{tab:complete_efficiency_results} reports the complete
efficiency measurements. The fixed 200-page latency sample contains
148,409 Full Index visual vectors in total. The NDCG@5 values are
evaluated on the complete ViDoRe v2 benchmark, whereas latency is
measured on the fixed 200-page sample.

At \(\gamma=0.10\), \method retains 93.8\% of the Full Index NDCG@5
while requiring \(1.109\times\) its total indexing latency,
corresponding to a 10.9\% indexing overhead. Relative to SAP, \method
improves NDCG@5 retention by 6.2 percentage points while increasing
total indexing latency by 4.811\,ms per page. Relative to Semantic
Clustering, \method improves NDCG@5 retention by 15.5 percentage points
while reducing total indexing latency by 19.641\,ms per page.
Semantic Clustering incurs the largest compression latency, whereas
1D-Pooling and DocPruner remain closest to the Full Index in total
indexing latency but preserve substantially less retrieval quality.
Overall, these results are consistent with
Figure~\ref{fig:efficiency_tradeoff}: \method provides the highest
retrieval fidelity among the evaluated compression methods while
incurring a moderate one-time increase in offline index construction
cost.

\end{document}